\documentclass[lettersize,journal]{IEEEtran}
\usepackage{amsmath,amsfonts}
\usepackage{algorithmic}
\usepackage{algorithm}
\usepackage{array}
\usepackage{textcomp}
\usepackage{stfloats}
\usepackage{url}
\usepackage{verbatim}
\usepackage{graphicx}
\usepackage{cite}
\usepackage{subfloat}
\usepackage{multirow} 
\usepackage{booktabs} 
\usepackage{overpic} 
\usepackage{subfigure} 
\usepackage{color}
\usepackage{xcolor}
\usepackage{makecell}
\usepackage{colortbl}  
\usepackage{rotating}
\usepackage{footnote}

\usepackage[colorlinks,linkcolor=red,anchorcolor=blue,citecolor=green]{hyperref} 
 
\newcommand{\etal}{{\textit{et al}}.\@ }
\definecolor{lightgrey}{RGB}{211,211,211}
\bstctlcite{IEEEexample:BSTcontrol}

\newcommand{\pink}[1]{\textcolor[rgb]{0,0,0} {#1}}
\newcommand{\orange}[1]{\textcolor[rgb]{0,0,0} {#1}}

\newcommand{\blue}[1]{\textcolor[rgb]{0,0,0} {#1}}
\newcommand{\purple}[1]{\textcolor[rgb]{0,0,0} {#1}}
\newcommand{\green}[1]{\textcolor[rgb]{0,0,0} {#1}}

\newcommand{\red}[1]{\textcolor[rgb]{0,0,0} {#1}}

\newcommand{\olive}[1]{\textcolor[rgb]{0,0,0} {#1}}
\newcommand{\rose}[1]{\textcolor[rgb]{0,0,0} {#1}}

\begin{document}

\title{\red{Region-Aware Portrait Retouching with Sparse Interactive Guidance}}
\author{Huimin Zeng, Jie Huang, Jiacheng Li, and Zhiwei Xiong

\thanks{This work was supported by the National Natural Science Foundation of China under Grants 62131003 and 62021001. (\textit{Corresponding author: Zhiwei Xiong.})  

The authors are with the Department of Electronic Engineer and Information Science, University of Science and Technology of China, Hefei, 230026, China (e-mail: zenghuimin@mail.ustc.edu.cn; hj0117@mail.ustc.edu.cn; jclee@mail.ustc.edu.cn; zwxiong@ustc.edu.cn).  

Code is released at \url{https://github.com/ZeldaM1/interactive_portrat_retouching}. } 

}

\markboth{Journal of \LaTeX\ Class Files,~Vol.~14, No.~8, August~2021}%
{Shell \MakeLowercase{\textit{et al.}}: A Sample Article Using IEEEtran.cls for IEEE Journals}


\maketitle

\begin{abstract}
Portrait retouching aims to improve the aesthetic quality of input portrait photos and especially requires human-region priority. \pink{The deep learning-based methods largely elevate the retouching efficiency and provide promising retouched results. However, existing portrait retouching methods focus on automatic retouching, which treats all human-regions equally and ignores users' preferences for specific individuals,} thus suffering from limited flexibility in interactive scenarios. In this work, we emphasize the importance of users' intents and explore the interactive portrait retouching task. Specifically, we propose a region-aware retouching framework with two branches: an automatic branch and an interactive branch.  \pink{The automatic branch involves an encoding-decoding process, which searches region candidates and performs automatic region-aware retouching without user guidance.   The interactive branch encodes sparse user guidance into a priority condition vector and modulates latent features with a region selection module to further emphasize the user-specified regions.  Experimental results show that our interactive branch effectively captures users' intents and generalizes well to unseen scenes with sparse user guidance, while our automatic branch also outperforms the state-of-the-art retouching methods due to improved region-awareness.}

\end{abstract}


\section{Introduction}\label{sec: intro}
\red{Aiming at promoting the flat-looking tone of raw inputs and emphasizing human regions, portrait retouching has a vast range of applications in practical scenarios such as advertisements, close-ups and group photos~\cite{zhao2021selective,liang2021ppr10k}.} Nevertheless, it is a challenge for amateurs to manually retouch a massive collection of portraits, which calls demands for automatic portrait retouching.  Recently, deep learning-based methods~\cite{liang2021ppr10k,zeng2020learning,he2020conditional,liang2021high} have been applied on this task.  \pink{However, they are not flexible enough \red{since} they ignore users' preferences for specific individuals and equally retouch all human-regions.} For instance, given a practical scenario of nonuniform illumination (e.g., the group photo shown in Fig.~\ref{fig:teaser}(a)), applying the automatic retouching without considering human-region priority and individual adaptivity inevitably leads to overexposure and oversaturation problems.  \pink{Therefore, it is essential to develop models that consider users' intents and emphasize the retouching of user-specified regions.}

\red{Portrait retouching is a special case of photo retouching. Current popular photo retouching datasets~\cite{mitfivek,hasinoff2016burst} ignore the demand for region awareness, and generate retouched ground truths by globally converting raw inputs with transformation curves.       Early methods~\cite{chen2018deep,kim2020pienet,deng2018aesthetic,zhang2019multiple,kim2020global,ni2020unpaired,ni2020towards} concentrate on generating more realistic results in an end-to-end manner.       Later methods~\cite{he2020conditional,liang2021high,moran2021curl,kim2021representative,zeng2020learning,wang2021real} focus on promoting the efficiency of photo retouching.      } Among these methods, 3D LUT~\cite{zeng2020learning} utilizes 3D lookup tables (LUTs) to achieve fast transformation.        \pink{Since these approaches are not designed for portrait scenarios}, 3D LUT HRP~\cite{liang2021ppr10k} proposes a spatial-adaptive portrait retouching dataset, namely, the PPR10K dataset, to meet the demand for portrait retouching and injects region-awareness into 3D LUT with a human-region priority strategy.
\pink{The methods mentioned above merely rely on input images to make predictions, albeit they behave in a fully automatic manner.    Therefore, they underestimate the impact of manual inputs, which strongly indicate users' preferences.}  \red{In addition, their performance may drop significantly in extreme cases that ambiguously define the portrait regions (e.g., low-resolution~\cite{ge2018low,gharbi2017deep}, low-light~\cite{he2020conditional,wang2018gladnet}, and occlusions~\cite{ge2020occluded,ge2017detecting}).}
\pink{This motivates us to develop a region-aware portrait retouching method that benefits from interactive user guidance to search retouching regions and meet various users' preferences.}

\red{A naive solution to achieve interactive retouching is shown in Fig.~\ref{fig:teaser}(b), which takes a cascaded strategy of ``interactive segmentation - local retouching - image harmonization".  The interactive segmentation network predicts the human-region mask according to user guidance.   Then the automatic retouching methods perform within the mask only. The harmonization network finally adjusts the blended image to make it realistic.     However, such a solution is incompetent to meet the following three major requirements for the interactive portrait retouching task:} (1) Human-region priority: Instead of retouching the whole image, portrait retouching is expected to be human-region aware, and the human region should be the focus of retouching.  (2) Naturalness: The emphasized and retouched regions are expected to be in harmony with other regions.  (3) Instance adaptivity: 3D LUT HRP~\cite{liang2021ppr10k} achieves human-region priority with semantic portrait-masks of each image, resulting in simultaneous retouching for all portraits.   We argue that the priority within human-regions should be ordered according to the users' intents.  \pink{Therefore, users are allowed to emphasize and embellish any portrait of individuals as demand.}

\red{To satisfy the above requirements, in this paper, we explore a new avenue toward interactive portrait retouching with sparse user guidance}.   
Specifically we design a region-aware retouching model with two functions (see Fig.~\ref{fig:teaser}(c)): (1) Given an input image to retouch, our model automatically retouches possible regions with region-awareness, while taking human-region priority and naturalness into consideration.
(2) With sparse user guidance, our model further emphasizes the retouching of user-specified regions, which accomplishes the goal of instance adaptivity. To implement the two functions above, we propose a unified framework with two branches for automatic and interactive region-aware retouching subtasks, respectively.

For the first subtask without user guidance, we construct an encoding-decoding based automatic region-aware retouching branch.  During encoding, a potential region extractor is adopted to search regions of interest and provide plausible region candidates, which are utilized by the feature fusion module to achieve region-awareness.  For the second subtask with user guidance, we introduce an interactive region-aware retouching branch to first encode the guidance into a priority condition vector and then emphasize the retouching of the user-specified regions by modulating region candidates under the condition vector with a region selection module. Additionally, to avoid the forgetting problem during learning automatic and interactive subtasks~\cite{serra2018overcoming}, we apply a stagewise training strategy to progressively integrate the user guidance.  We show that the proposed method significantly promotes the quality of input images, and more importantly, it provides flexibility for users to retouch portraits interactively.

In conclusion, our contributions are summarized as follows:
\begin{itemize}
\item By introducing sparse user guidance into the retouching process, \red{we emphasize the benefit of users' intents and explore a new avenue toward the interactive portrait retouching task.}
\item We propose a unified framework consisting of automatic and interactive region-aware retouching branches.  \red{Based on the automatic branch, the interactive branch further incorporates user guidance with extracted semantics through a region selection module to achieve instance adaptivity.}
\item \pink{Experimental results show the ability of our interactive retouching branch to effectively capture users’ intents and the generalization ability on non-portrait-dominated scenarios, as well as the state-of-the-art performance of our automatic retouching branch.}
\end{itemize}

\section{Related work}
\subsection{Automatic Photo Retouching}
Since photo retouching methods can also be applied to the portrait retouching task, we first introduce the photo retouching methods and then introduce the portrait-specific approach. We group current photo retouching methods into image-to-image translation methods and operator prediction methods.

\textbf{Image-to-image Translation Methods.}
Image-to-image translation models perform the photo retouching task in an end-to-end manner. Ignatov~\etal~\cite{ignatov2017dslr} propose to obtain retouched results with residual learning, making progress in both contrast enhancement and edge maintenance. Unpaired learning methods~\cite{chen2018deep,deng2018aesthetic,zhang2019multiple,kim2020global,ni2020unpaired,ni2020towards} explore weakly supervised retouching with generative adversarial networks~\cite{goodfellow2014generative}. LPTN~\cite{liang2021high} decomposes inputs with the Laplacian pyramid and performs translation on low-frequency components while preserving high-frequency with a progressive masking strategy. StarEnhancer~\cite{song2021starenhancer} introduces multiple-style enhancement with the ability to transform inputs to an unseen style. CSRNet~\cite{he2020conditional} designs a lightweight framework containing a base network and a conditioning network for extracting global features and performing photo retouching, respectively. Pienet~\cite{kim2020pienet} constructs preference vectors with metric learning and adaptively enhances images according to user-provided preferable styles.

\textbf{Operator Prediction Methods.}
DeepLPF~\cite{moran2020deeplpf} regresses the parameters of spatially localized filters and automatically applies those filters to enhance inputs.   
HDRNet~\cite{gharbi2017deep} obtains transformations on the low-resolution input and applies upsampled transformations to the full-resolution image with bilateral grid processing.  
Enhancement curve learning methods~\cite{chai2020supervised,li2020flexible,moran2021curl} estimate retouching curves to tone global properties of inputs rather than directly mapping.  
RCTNet~\cite{kim2021representative} first estimates the transformation of the representative colors and then enhances inputs according to the similarity between inputs and representative colors. 
3D LUT based methods~\cite{zeng2020learning,wang2021real} utilize variants of 3D lookup tables (3D LUTs) with deep learning, achieving real-time and flexible photo enhancement performance. 

\blue{In contrast to most of the methods above, 3D LUT HRP~\cite{liang2021ppr10k} focuses on portrait retouching and achieves region awareness through the human-region priority strategy.} Our work distinguishes from 3D LUT HRP~\cite{liang2021ppr10k} by emphasizing the significance of the user guidance and investigating interactive region-aware portrait retouching, whcih provides the flexibility to retouch different instances according to users' intents.

\subsection{Interactive Image Editing}
Current interactive image editing works mainly concentrate on image synthesis, style transfer and colorization tasks. \blue{Depending on the types of interactive guidance, we group the interactive image editing methods into attribute-based methods and sketch/click-based methods.}

\blue{\textbf{Attribute-based Methods.}
AnycostGAN~\cite{lin2021anycost} runs with a mini-generator and outputs results of flexible resolution for interactive face attribute editing.
SEAN~\cite{zhu2020sean} presents semantic region-adaptive normalization to facilitate image editing control.  
NSP~\cite{virtusio2021neural} considers different attributes in the single style image as the anchor styles to provide visual guidance for users.
EditGAN~\cite{ling2021editgan} optimizes conditional latent vectors according to interactively modified semantic masks and achieves semantic image edits.  
Wang \etal~\cite{wang2018high} propose a multiscale generator for high-resolution image synthesis and interactively manipulate with additional user-given features.}

\begin{figure*}[t]
\centering
\vspace{-0.1in}\includegraphics[width=1\linewidth]{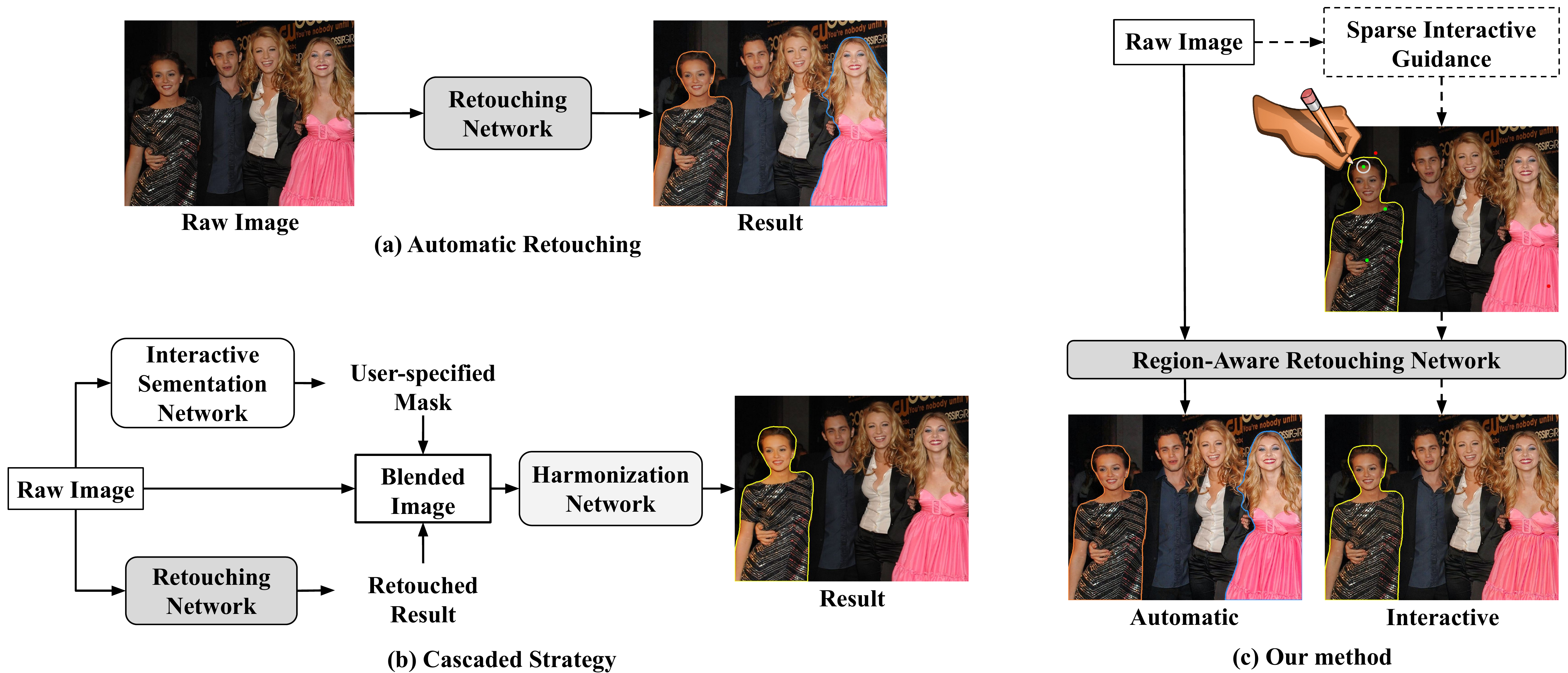}\vspace{-6pt}
\caption{Comparison between the proposed region-aware retouching method and existing retouching methods.   (a) Automatic retouching, which retouches the input raw image in an end-to-end manner.   Here we provide the automatic retouching result of 3D LUT~\cite{zeng2020learning}.   (b) The cascaded strategy, which adopts a naive solution of \textit{``interactive segmentation - local retouching - image harmonization"} to achieve interactive retouching.   Here we adopt RITM-H18~\cite{sofiiuk2021reviving}, 3D LUT~\cite{zeng2020learning} and DHT~\cite{guo2021image} to complete the steps mentioned above.    (c) Our method.  The interactively emphasized/non-emphasized retouching regions are indicated by green/red dots.  As can be seen, automatic retouching treats individuals equally and retouches multiple instances (marked with orange and blue curves, respectively) simultaneously, leading to local overexposure (e.g., the girl dressed in pink). Meanwhile, the cascaded strategy introduces oversaturation and unnaturalness for the locally retouched region (marked with a yellow curve).   \red{In contrast, our automatic branch keeps a balance between different instances, while our interactive branch pays special attention to the user-specified region (marked with a yellow curve).}}
\label{fig:teaser} 
\end{figure*}

\blue{\textbf{Sketch/click-based Methods.}
Huang~\etal~\cite{huang2021multi} explore the sketch-to-image translation task with multi-level user guidance. 
Given body skeletons/landmarks, C2GAN~\cite{tang2021total} introduces a cycle in cycle generative adversarial network to boost the gesture generation process. 
Li~\etal~\cite{li2020staged} employ a conditional generative adversarial network for class-wise representation extraction and sketch-to-image synthesis.
ClickMatter~\cite{gao2022clicking} employs user-given clicks to generate semantic-aware localization maps for the human parsing task. 
Zhang~\etal~\cite{zhang2017real} propose fusing low-level and high-level semantics to propagate user-given clicks and generate possible colors. 
}

\red{It is worth mentioning that although Deepfake~\cite{ge2022deepfake,9270435} also offers an interactive solution with generative models, it concentrates on replacing target faces with synthesized ones, while photo retouching stylizes the entire raw inputs rather than local faces. Besides, Deepfake and the above methods focus on high-level attribute editing, which generate results with dramatically different content from inputs. In contrast, our work requires high fidelity of outputs so that only low- level changes such as brightness, contrast, and saturation are involved. Due to the simple and user-friendly characteristics of clicks, we adopt clicks as the interactive guidance in this work.}

\section{Method}
\subsection{Overview}
\begin{figure*}[t]
  \centering
    \vspace{-0.1in}\includegraphics[width=0.97\linewidth]{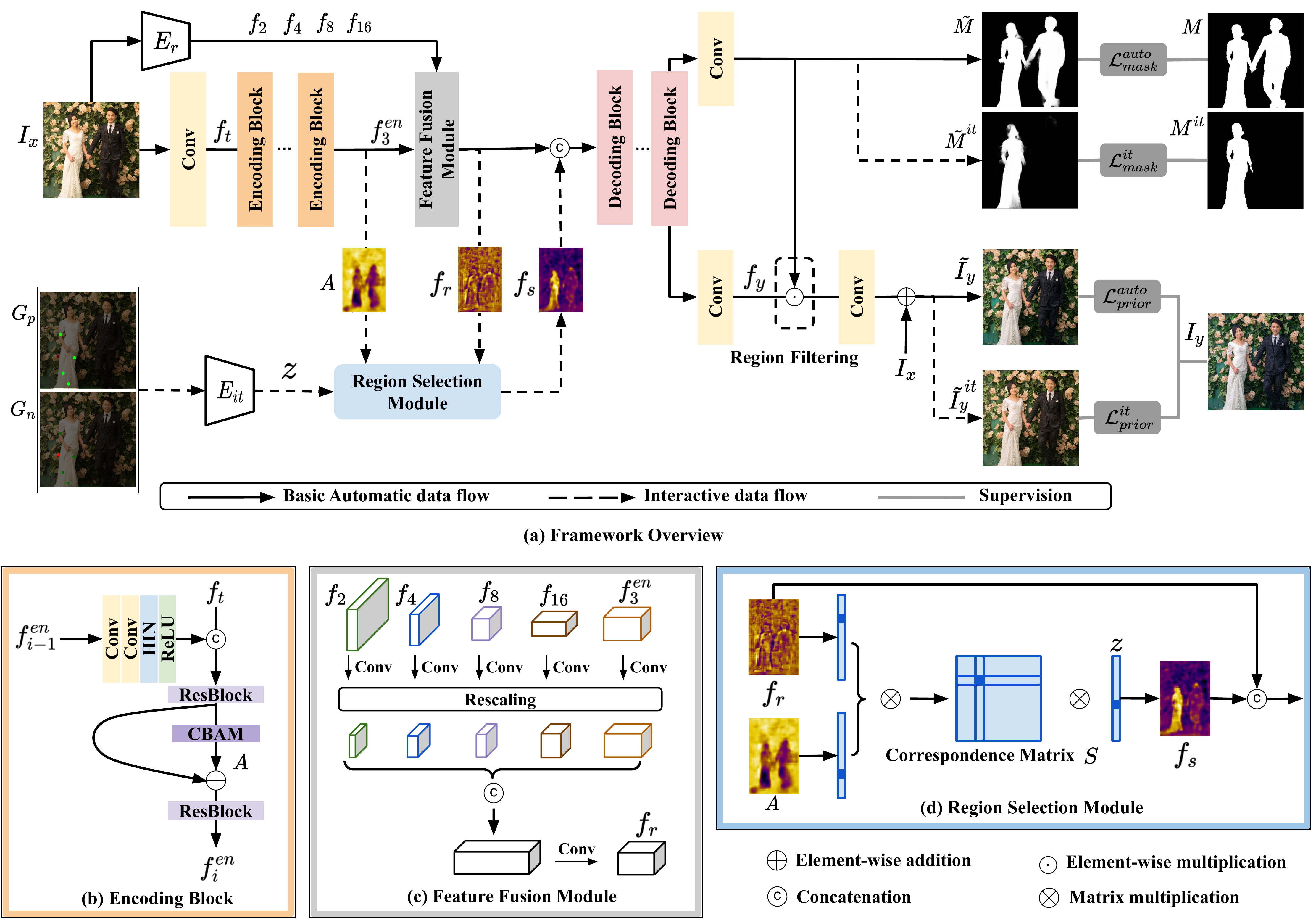}\vspace{-4pt}
\caption{
(a) Overview of the proposed framework, which consists of an automatic region-aware retouching branch and an interactive region-aware retouching branch (denoted as solid lines and dashed lines, respectively).            The automatic branch (solid lines) involves an image encoding and decoding process to give retouched $\tilde{I}_y$ from input $I_x$.           The interactive region-aware retouching branch (dashed lines) further encodes user guidance [$G_p$, $G_n$] into priority condition vector $z$ with encoder $E_{it}$.         The region-aware latent feature $f_r$ is modulated by $z$ in the proposed region selection module to achieve instance adaptivity.
(b) Encoding block, which extracts image semantics and generates human-region attention $A$.       HIN denotes the half instance norm~\cite{chen2021hinet}, and CBAM denotes the convolutional block attention module~\cite{woo2018cbam}.            The encoding block turns into a decoding block by replacing the first convolution layer with a deconvolution layer.
(c) Feature fusion module, which integrates multiscale ROI features [$f_2$, $f_4$, $f_8$, $f_{16}$] with image semantics $f_3^{en}$ and obtains region-aware latent feature $f_r$.
(d) Region selection module, which specifies the emphasized region with condition vector $z$ by exploring the semantic correspondence $\mathcal{S}$ between the human-region attention $A$ and region-aware feature $f_r$.} 
  \label{Framework overview}
\end{figure*}

As illustrated in Fig.~\ref{Framework overview}(a), our unified framework consists of a basic automatic region-aware retouching branch and an interactive region-aware retouching branch, which are denoted by the solid and dashed lines, respectively. \red{Verbal representations of symbols in Fig.~\ref{Framework overview} are provided in TABLE~\ref{tab:acronyms} to facilitate reading.} Given an input image $I_x$ under the user-guidance agnostic situation, we seek to predict the retouched result $\tilde{I}_y$ with the automatic region-aware retouching branch, which contains a basic image encoding and decoding process. \purple{To provide flexibility to retouch different instances according to users' intents, we introduce sparse positive clicks and negative clicks (denoted as green dots in $G_p$ and red dots in $G_n$) to indicate emphasized and non-emphasized retouching regions, respectively}, and then modulate region candidates with the interactive region-aware retouching branch. Detailed descriptions for the automatic region-aware retouching branch and the interactive region-aware retouching branch are stated in Sec.~\ref{automatic retouching} and Sec.~\ref{interactive retouching}.

\begin{table}[t]
\centering
\red{
\caption{Verbal representations of symbols in Fig.~\ref{Framework overview}.}
\label{tab:acronyms}
\setlength{\tabcolsep}{2.8mm}{
\vspace{-4pt}\begin{tabular}{c|l}
\toprule
Symbol & \makecell[c]{Verbal Representation}  \\
\hline
$I_x$  & Input image \\
$I_y$ & Ground truth image \\
$\tilde{I}_y$  & Retouched result of the automatic branch \\
$\tilde{I}_y^{it}$ & Retouched result of the interactive branch \\
\hline
$G_p$  & Positive user guidance map \\
$G_n$ &  Negative user guidance map\\
\hline
$M$  & Ground truth  human-region mask \\
$\tilde{M}$ & Predicted human-region mask \\
$M^{it}$  & Ground truth human instance mask \\
$\tilde{M}^{it}$ &  Predicted human instance mask \\
\hline
$f_t$  & Textural feature \\
$f_3^{en}$ & Encoded feature of the $3^{nd}$ encoding block \\
$f_r$  & Region-aware feature \\
$f_s$ & Region-specified feature \\
$f_y$  & Decoded feature  \\
$f_2$ & Feature from the \textit{conv1} layer of $E_r$\\
$f_4$  & Feature from the \textit{conv2\_x} layer of $E_r$\\
$f_8$  & Feature from the \textit{conv3\_x} layer of $E_r$\\
$f_{16}$  & Feature from the \textit{conv4\_x} layer of $E_r$\\
\hline
$z$ & Priority condition vector \\
$A$  & Human-region attention \\
$S$ &  Correspondence matrix \\
\hline
$E_r$  & Potential region extractor \\
$E_{it}$  & Interactive representation encoder \\
    \bottomrule
  \end{tabular}\vspace{-8pt}
}
}
\end{table}

\subsection{Automatic Region-aware Retouching Branch}
\label{automatic retouching}
The automatic retouching branch (solid lines) is displayed in Fig.~\ref{Framework overview}(a) as an encoding and decoding process. Given an input image $I_x$, this branch aims to predict the region-aware result $\tilde{I}_y$ automatically and is supposed to meet two requirements: human-region priority and naturalness. We introduce the encoding and decoding process as follows.  

\subsubsection{Encoding} As shown in Fig.~\ref{Framework overview}(a), we divide the encoding process into three parts: (1) Encoding blocks; (2) Potential region extractor $E_r$; and (3) Feature fusion module.  
Given input image $I_x$, we pass $I_x$ into sequential encoding blocks and potential region extractor $E_r$ for image semantics extraction and exploring regions of interest (ROI), respectively. The extracted image semantic $f_3^{en}$ and multiscale ROI features [$f_2$, $f_4$, $f_8$, $f_{16}$] are then fused with the feature fusion module to further locate the target region and obtain the region-aware latent feature $f_r$.  

\textbf{Encoding Blocks.} To extract semantics from the inputs, we apply the encoding block demonstrated in Fig.~\ref{Framework overview}(b), which is constructed with convolution layers and CBAM-integrated ResBlock~\cite{woo2018cbam}.  For the $i^{th}$ encoding block, its inputs are encoded feature $f_{i-1}^{en}$ from the previous encoding block and the textural feature $f_t$ from the very first convolution layer of the automatic retouching branch.  The number of encoding blocks is set to 3 in practice.

\begin{figure}[t]
\centering
\vspace{-4pt}\hspace{-2pt}\includegraphics[width=63pt]{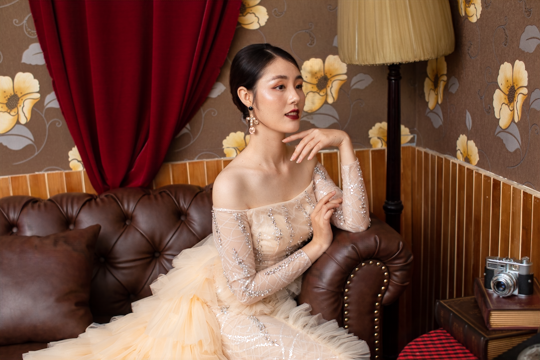}\hspace{-2pt} 
\includegraphics[width=63pt]{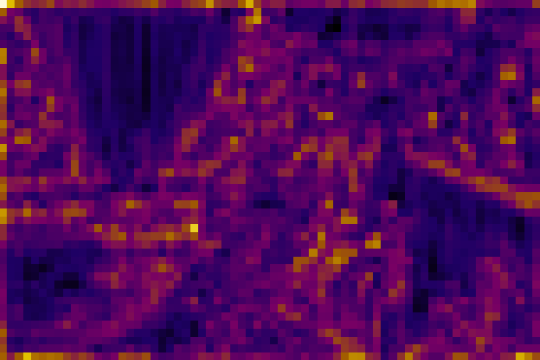}\hspace{-2pt} 
\includegraphics[width=63pt]{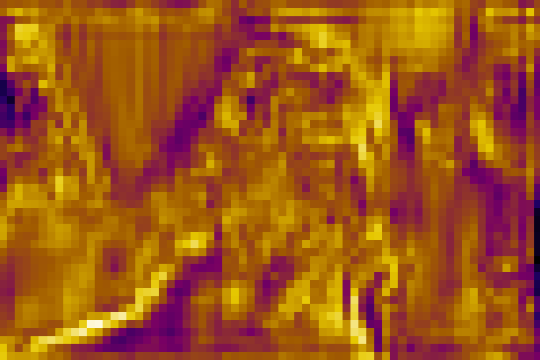}\hspace{-2pt} 
\includegraphics[width=63pt]{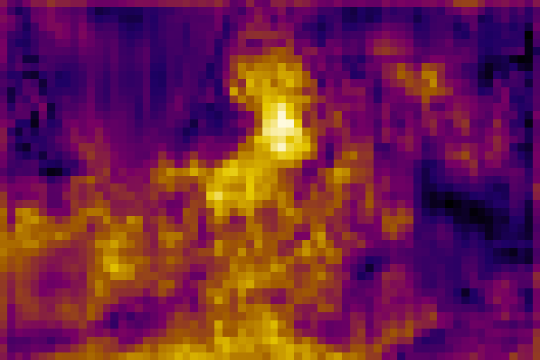}\hspace{-4pt}\vspace{-3pt}

\hspace{-2pt}\subfigure[Input]{\includegraphics[width=63pt]{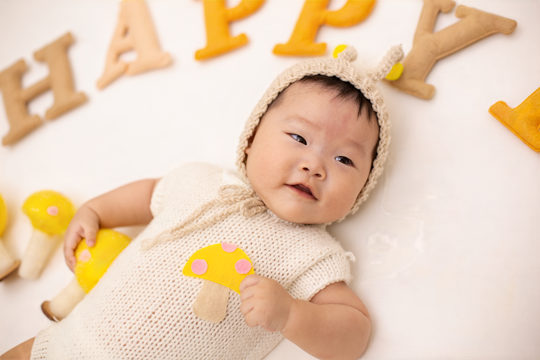}}\hspace{-2pt} 
\subfigure[$f_8$]{\includegraphics[width=63pt]{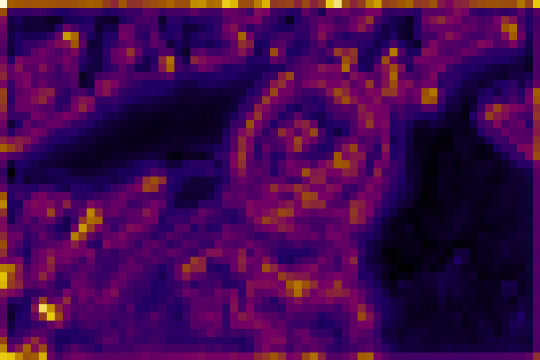}}\hspace{-2pt} 
\subfigure[$f_3^{en}$]{\includegraphics[width=63pt]{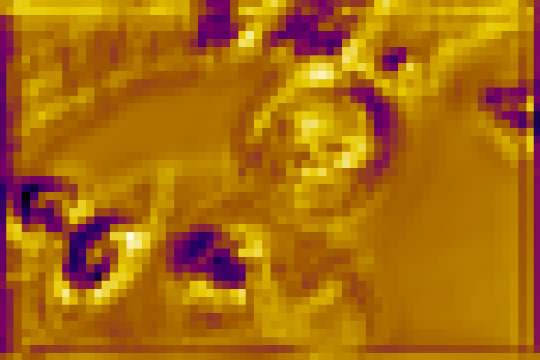}}\hspace{-2pt} 
\subfigure[$f_r$]{\includegraphics[width=63pt]{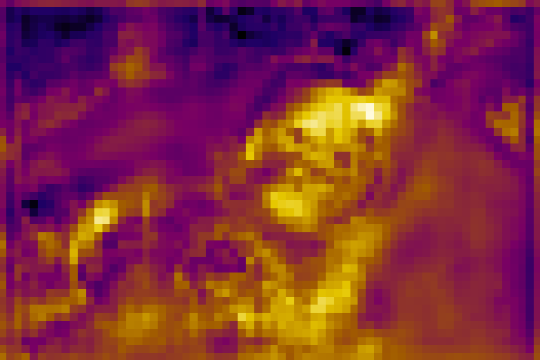}}\hspace{-4pt}

\vspace{-2pt}\caption{Visualization of the feature fusion module in the automatic retouching branch. The regions of interest $f_8$ mainly searches plausible retouching regions containing both human-regions and background. The region-aware feature $f_r$, which locates accurate human-regions, can be effectively established from the image semantic feature $f_3^{en}$.}\vspace{-8pt}

\label{fig: feature fusion module}
\end{figure}

\textbf{Potential Region Extractor $\boldsymbol{E_r}$.} To search regions of interest and obtain region candidates, ResNet-18~\cite{he2016identity} is adopted as the potential region extractor $E_r$. To better capture the global context, we extract multiscale features [$f_2$, $f_4$, $f_8$, $f_{16}$] from the \textit{conv1, conv2\_x, conv3\_x} and \textit{conv4\_x} layers, respectively. Subscripts (e.g.,2 and 4) indicate the downscaling factor compared with input image $I_x$. We provide visualization of $f_8$ in Fig.~\ref{fig: feature fusion module}(b), which shows regions of interest containing both the human-regions and the background (e.g., the baby and the text).

\textbf{Feature Fusion Module.} 
To accurately locate and emphasize human-related regions, we design the feature fusion module shown in Fig.~\ref{Framework overview}(c),  which fuses image semantics and region candidates adequately. \olive{As demonstrated in Fig.~\ref{Framework overview}(c), multiscale ROI features [$f_2$, $f_4$, $f_8$, $f_{16}$] and encoded image semantic $f_3^{en}$ are first passed through separate convolution layers and then rescaled to the same resolution with bilinear interpolation. These rescaled features are concatenated along the channel dimension and projected with a 3x3 convolution layer to obtain the region-aware latent feature $f_r$. } As shown in Fig.~\ref{fig: feature fusion module}, the region of interest $f_8$ extracted by the potential region extractor $E_r$ contains abundant possible regions (e.g., human-regions and background), and the proposed feature fusion module effectively locates human-related regions within $f_3^{en}$ and predicts the region-aware latent feature $f_r$.

\subsubsection{Decoding} \label{decoder}
Our decoding block is implemented by replacing the first convolution layer of the encoding block in Fig.~\ref{Framework overview}(b) with a deconvolution layer. In this subsection, we mainly introduce the following strategies that directly satisfy the human-region priority and naturalness requirements: (1) Region filtering design and (2) Residual learning.

\textbf{Region Filtering.}
In contrast to the feature fusion module which helps to locate human-related regions, the region filtering strategy directly filters out human-irrelevant regions. We achieve region filtering with a modified feature masking mechanism~\cite{featuremasking}. As demonstrated in Fig.~\ref{Framework overview}(a), the human-region mask $\tilde{M}$ or human instance mask $\tilde{M}^{it}$ is obtained by projecting the human-region attention $A$ from the last decoding block with a 3x3 convolution layer, depending on whether facing interactive scenarios. By applying element-wise multiplication between the decoded features $f_y$ and human-region mask $\tilde{M}$ (or human instance mask $\tilde{M}^{it}$), we significantly dispose of the human-irrelevant region and guarantee the human-region priority.

\textbf{Residual Learning.} Since region-aware retouching especially requires local adjustment and maintaining fidelity, we employ skip connection~\cite{he2016deep,ignatov2017dslr} at the end of the retouching network to ease the retouching process and preserve the naturalness.

\subsection{Interactive Region-aware Retouching Branch}
\label{interactive retouching}
Since portrait retouching is a highly user-dependent subtask, while automatically retouched results may fail to meet different users’ favor, we believe there is a need to grant access for users to retouch images according to their preferences and achieve instance adaptivity. In the interactive retouching task, the core is how to associate sparse user guidance with currently extracted image semantics from the automatic region-aware retouching branch. \pink{To address this issue, we design the region selection module shown in Fig.~\ref{Framework overview}(d), which retrieves user-specified regions from extracted semantics under user guidance with a correspondence matrix.}  In this subsection, we first introduce the interactive retouching branch displayed in Sec.~\ref{sec:Network Architecture} and then elaborate the proposed region selection module in Sec.~\ref{sec:Region Selection Module}.

\subsubsection{Network Architecture}\label{sec:Network Architecture}
As shown in Fig.~\ref{Framework overview}(a), based on the automatic region-aware retouching branch, we introduce an additional interactive branch (dashed lines) for user-customized retouching, which models interactive retouching as a latent feature editing task toward region-aware feature $f_r$ with sparse user guidance. \green{Given an input image $I_x$, users are expected to specify the emphasized/non-emphasized retouching regions with positive/negative clicks (denoted as green/red dots in Fig.~\ref{Framework overview}(a)). In practice, the positive/negative clicks are first converted into binary user guidance map $G_p$ and $G_n$.} The concatenation of $G_p$ and $G_n$ is then encoded with sequential convolution layers (denoted as $E_{it}$) and reshaped into the priority condition vector $z$. The region-aware latent feature $f_r$ is further modulated by the priority condition vector $z$ with our region selection module to achieve user-customized retouching.

\subsubsection{Region Selection Module}\label{sec:Region Selection Module} 
To specify the emphasized region with condition vector $z$, we design the region selection module shown in Fig.~\ref{Framework overview}(d). Our basic idea is to 
consider the human-region attention $A$ and region-aware feature $f_r$ as a \textit{key-value} pair in the latent space, from which the region-specified feature $f_s$ is retrieved under the priority condition vector $z$. Given the \textit{query} priority condition vector $z$, we compare it with the \textit{key} human-region attention $A$ which contains highly-relevant human-region sources, and retrieve the corresponding \textit{value} region-specified feature $f_s$. We divide the working process of the region selection module into two steps: correspondence matrix calculation and user-input propagation.

\begin{figure}[t]\vspace{-4pt}
\hspace{-1pt}\includegraphics[width=41pt]{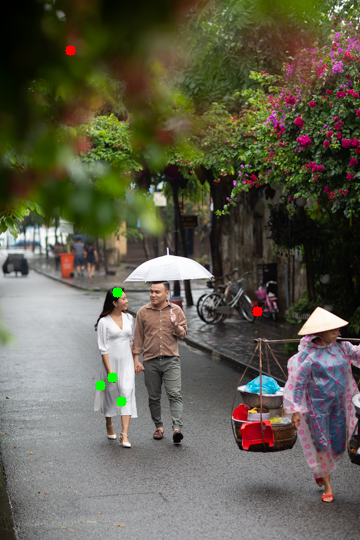}\hspace{-1pt} 
\includegraphics[width=41pt]{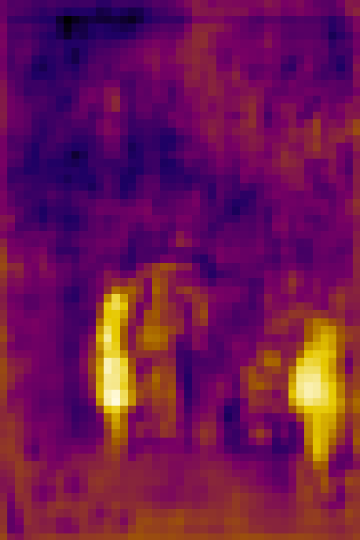}\hspace{-1pt} 
\includegraphics[width=41pt]{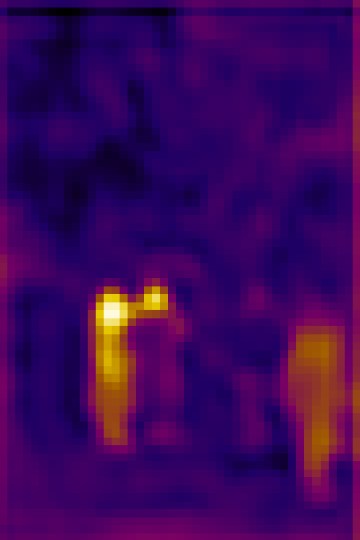}\hspace{-1pt} 
\includegraphics[width=41pt]{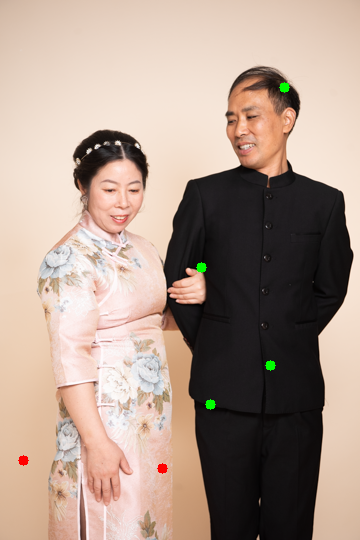}\hspace{-1pt} 
\includegraphics[width=41pt]{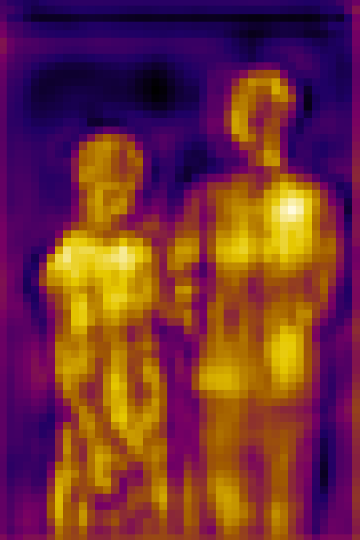}\hspace{-1pt} 
\includegraphics[width=41pt]{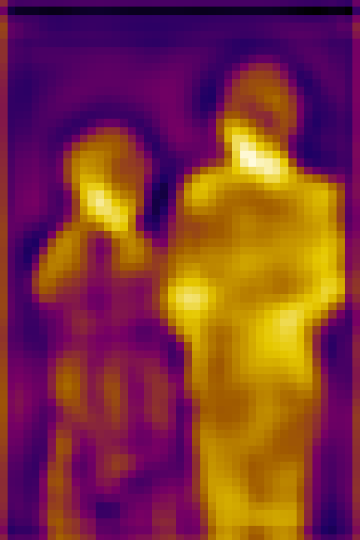}\vspace{-3.5pt}

\hspace{-1pt}\subfigure[Input]{\includegraphics[width=41pt]{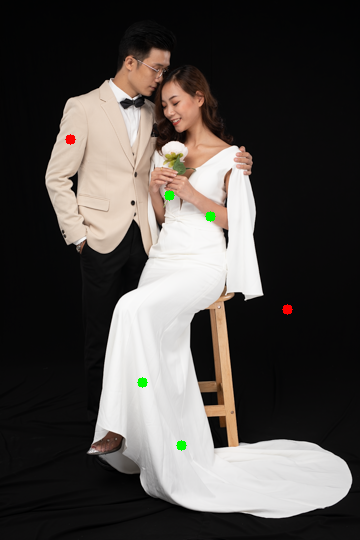}}\hspace{-1pt} 
\subfigure[$f_r$]{\includegraphics[width=41pt]{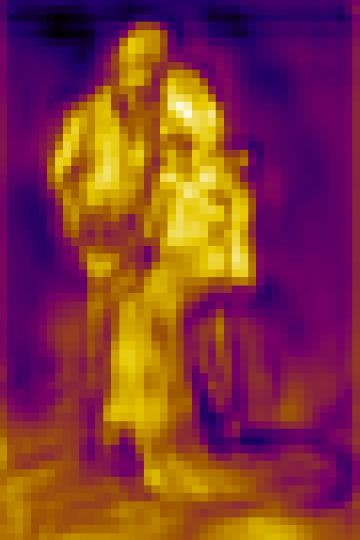}}\hspace{-1pt} 
\subfigure[$f_s$]{\includegraphics[width=41pt]{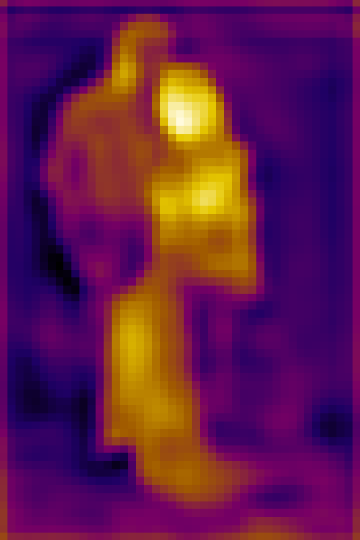}}\hspace{-1pt} 
\subfigure[Input]{\includegraphics[width=41pt]{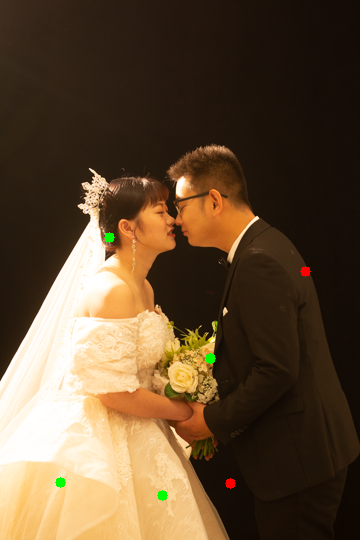}}\hspace{-1pt} 
\subfigure[$f_r$]{\includegraphics[width=41pt]{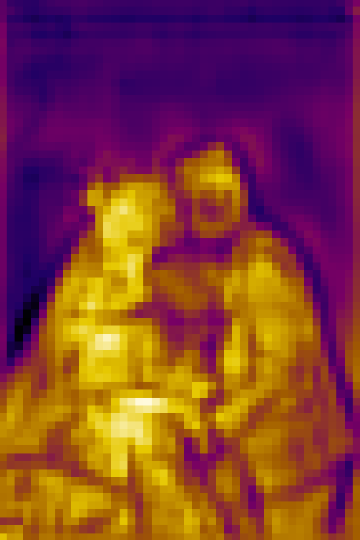}}\hspace{-1pt} 
\subfigure[$f_s$]{\includegraphics[width=41pt]{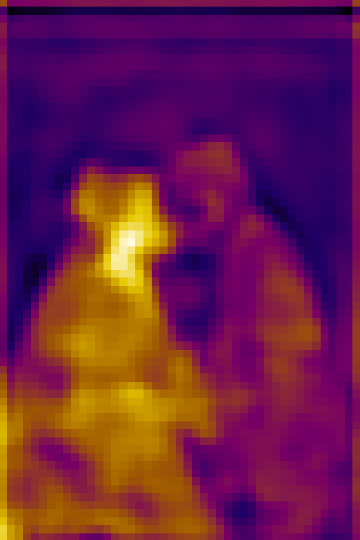}} \\

\vspace{-8pt}\caption{The influence of sparse user guidance in the proposed region selection module. The region selection module effectively specifies the retouching region from the region-aware latent feature $f_r$ under sparse user guidance, obtaining the region-specified feature $f_s$.}\vspace{-8pt}  
\label{fea}
\end{figure}

\textbf{Correspondence Matrix Calculation.} 
We first perform a pairwise correspondence calculation to establish the \textit{key-value} pair of the human-region attention $A$ and region-aware feature $f_r$. Given region-aware feature $f_r \in \mathbb{R}^{C \times  H \times  W}$ and its corresponding human-region attention $A \in \mathbb{R}^{C \times H \times W}$.  We first reshape $f_r$ and $A$ to vectors $\hat{f}_r\in \mathbb{R}^{C \times HW}$ and $\hat{A} \in \mathbb{R}^{C \times HW}$ respectively. A dense correspondence matrix $\mathcal{S} \in \mathbb{R}^{HW\times HW}$ is obtained by computing a pairwise correlation between $\hat{f}_r$ and $\hat{A}$ as follows:
\begin{equation}
\mathcal{S}(i,j)=\frac{\hat{f}_r(i)^{T} \hat{A}(j)}{\left\|\hat{f}_r(i)\right\|\left\|\hat{A}(j)\right\|}.
\end{equation}

\textbf{User-Input Propagation.} 
Given the \textit{query} priority condition vector $z \in \mathbb{R}^{HW \times C}$, we retrieve the \textit{value} region-specified feature $f_s$ from the established \textit{key-value} pair by selecting the most correlated information between the \textit{query} $z$ and the \textit{key} $A$. The propagation process is as follows:

\vspace{-1pt}\begin{equation}
\hat{f}_{s}(i)=\sum_{j} \operatorname{Softmax}(\mathcal{S}(i, j) / \tau) \cdot z(j), 
\end{equation}
where $\tau$ denotes the scaling factor that controls the sharpness of the softmax and is set to 1 by default. Vector $\hat{f}_{s} \in \mathbb{R}^{HW \times C}$ is later reshaped into the region-specified feature map $f_s \in \mathbb{R}^{C \times H \times W}$ and concatenated with $f_r$ to go through the decoding process.  We provide visualizations of features in region selection module in Fig.~\ref{fea}, which clearly shows how the sparse user guidance helps to emphasize retouching regions and obtain region-specified feature $f_s$ from region-aware feature $f_r$.

\subsection{Training}
\begin{figure}[t]
\centering
\vspace{-10pt}\includegraphics[width=255pt]{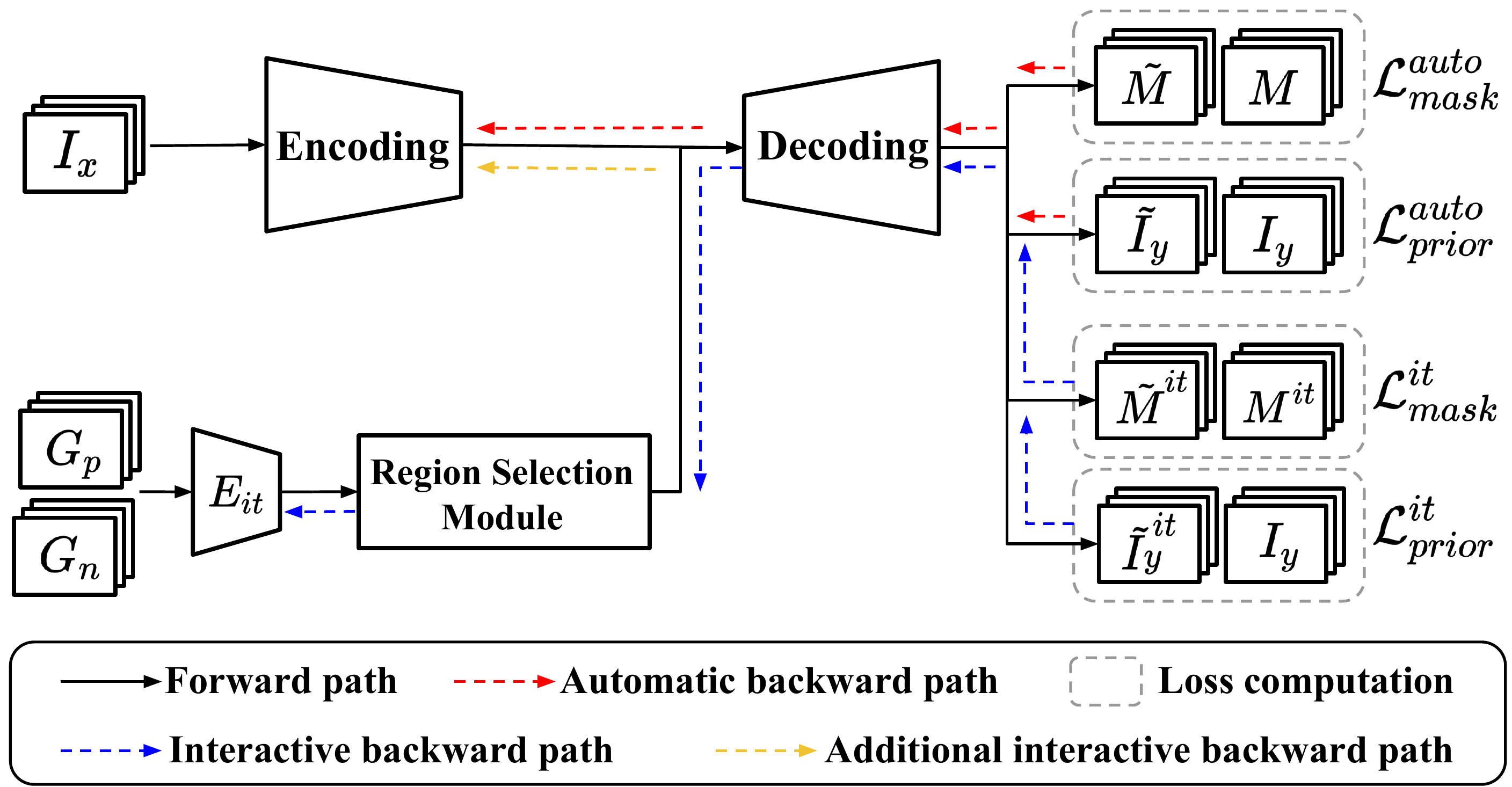}\vspace{-4pt}
\caption{Illustration of the stagewise training strategy, where the red dashed lines and blue dashed lines stand for the backward propagation path of automatic retouching training and interactive training respectively. For the joint training of two branches, the backward propagation path is either red dashed lines or blue-yellow dashed lines, depending on the training data fed in.
}\vspace{-8pt}
\label{fig: training strategy}
\end{figure}

As demonstrated in Fig.~\ref{fig: training strategy}, the proposed stagewise training strategy consists of three stages: 1) Automatic retouching branch training; 2) Interactive retouching branch training; and 3) Joint training of two branches. Based on the preliminary region-aware information learned in the first stage, the interactive retouching ability is further obtained with the aid of user guidance.

The training loss for each stage consists of a human-region priority term $\mathcal{L}_{priority}$ and a human-region mask term $\mathcal{L}_{mask}$, 
\begin{equation}
\mathcal{L}_{total}=\lambda_p\mathcal{L}_{priority} +\lambda_m\mathcal{L}_{mask},
\end{equation}
where $\lambda_p$ and $\lambda_m$ denote the coefficients of human-region priority loss $\mathcal{L}_{priority}$ and human-region mask loss $\mathcal{L}_{mask}$, respectively. We set $\lambda_p$ and $\lambda_m$ to 1 in practice.

\textbf{Automatic Retouching Training.}
\label{HRP strategy}
 We first train the automatic retouching branch while freezing the interactive branch, and the backward propagation path is demonstrated with red dashed lines in Fig.~\ref{fig: training strategy}. Following the human-region priority (HRP) strategy~\cite{liang2021ppr10k}, we employ the human-region weighted L1 loss on retouched results to emphasize the human-region,
\begin{equation}
\label{retouching}
\mathcal{L}_{priority}^{auto}=\left\|\boldsymbol{W} \odot \boldsymbol{\tilde{I}_y} - \boldsymbol{W} \odot \boldsymbol{I_y}\right\|,
\end{equation}
where $\boldsymbol{\tilde{I}_y}$, $\boldsymbol{I_y}$ and $\boldsymbol{W}$ denote the automatically retouched result, the ground truth image and the weighting matrix, respectively. 
We set $w_{ij}= 5$ for human-regions while $w_{ij}= 1$ for other regions.  Since we apply the region filtering strategy to specify the human-region and dispose of human-irrelevant regions, the following BCE loss is utilized to explicitly constrain the human-region mask,
\begin{equation}
\label{mask}
\mathcal{L}_{mask}^{auto} =-(\boldsymbol{M} \log (\boldsymbol{\tilde{M}})+(1-\boldsymbol{M}) \log (1-\boldsymbol{\tilde{M}})),
\end{equation}
where $\boldsymbol{\tilde{M}}$ denotes the predicted human-region mask and $\boldsymbol{M}$ represents the ground truth human-region mask.

\textbf{Interactive Retouching Training.}
Based on the automatic retouching branch, we train the interactive retouching branch while freezing the encoding process of the automatic retouching branch. The backward propagation path of interactive retouching is illustrated with blue dashed lines in Fig.~\ref{fig: training strategy}. For this branch, we adopt the same constraint term as Eq.~\ref{retouching} and Eq.~\ref{mask} while replacing human-region mask $\boldsymbol{M}$ with human instance mask $\boldsymbol{M}^{it}$. The specific loss functions are shown as follows:
\begin{equation}
\label{inter retouching}
\mathcal{L}_{priority}^{it}=\left\|\boldsymbol{W} \odot \boldsymbol{\tilde{I}_y^{it}} - \boldsymbol{W} \odot \boldsymbol{I_y}\right\|,
\end{equation}
where $\boldsymbol{\tilde{I}_y^{it}}$ denotes the interactive retouched results and $\boldsymbol{W}$ is defined in Eq.~\ref{retouching}. We set $w_{ij}= 5$ for human instance regions \and{and} $w_{ij}= 1$ for others.
Similarly, the mask constraint term is expressed as follows:
\begin{equation}
\label{inter mask}
\mathcal{L}_{mask}^{it} =-(\boldsymbol{M}^{it} \log (\boldsymbol{\tilde{M}}^{it})+(1-\boldsymbol{M}^{it}) \log (1-\boldsymbol{\tilde{M}}^{it})),
\end{equation}
 where $\boldsymbol{\tilde{M}}^{it}$ denotes the predicted human instance mask and $\boldsymbol{M}^{it}$ denotes the ground truth human instance mask.

\textbf{Joint Training of Two Branches.}
After separately training the automatic retouching branch and interactive retouching branch, we jointly train the two branches to bridge the gap between subtasks~\cite{serra2018overcoming}. For each iteration, the training data are randomly chosen from the automatic dataset and interactive dataset. As shown in Fig.~\ref{fig: training strategy}, during this stage, the yellow dashed line is merged with the blue dashed lines as the backward propagation path of the interactive branch. Thus, depending on the type of input data, the backward  propagation path is either red dashed lines or blue-yellow dashed lines, and the two branches are then optimized with corresponding loss functions Eq.~\ref{retouching}-\ref{mask} and Eq.~\ref{inter retouching}-\ref{inter mask}.

\begin{table*}[t] 
  \centering
  \vspace{-8pt}\caption{Comparison with state-of-the-art methods. Our automatic branch outperforms baseline methods in terms of almost all metrics, especially in human-centered (HC) ones.} 
  \label{quantitative conmparison}
    \setlength{\tabcolsep}{2.5mm}{
  \begin{tabular}{c | c | c c c c c | c c }
    \toprule
   Dataset & Method &PSNR $\uparrow$ &  $\Delta E_{a b} \downarrow$ & SSIM $\uparrow$ & MS-SSIM $\uparrow$  & NIQE $\downarrow$  & PSNR\small{$^{HC}$} $\uparrow$ & $\Delta E_{a b}$\small{$^{HC}$} $\downarrow$ \\  
  
   \hline
     \multirow{7}{*}{PPR10K-a} & UEGAN~\cite{ni2020towards}  & 14.47 & 	24.52 &  0.5409	 & 0.7667 &    6.9312 & 17.67 & 16.14  \\ 
     & HDRNet~\cite{gharbi2017deep}   &   22.79 &10.23 & 	0.9189 & 	0.9628   &3.9791  & 26.08 & 	6.68\\
      & CSRNet~\cite{he2020conditional}   & 22.44 & 10.00& 	0.9245& 	0.9643  & 4.0549  & 25.74 & 	6.50\\

      & LPTN~\cite{liang2021high}   &  21.62 & 10.30& 	0.8992	& 0.9372 & 3.9795   & 25.09	 & 6.65\\
      & 3D LUT~\cite{zeng2020learning}   &    25.81 & 7.02& 	0.9501	& 0.9779&   4.0009 & 29.08 & 	4.55 \\ %

     &  3D LUT HRP~\cite{liang2021ppr10k}   &    25.98 &6.76 & 0.9523	&  \textbf{0.9805 }&  3.9595  & 29.29 & 	4.38\\ %
     & \textbf{Ours} & \textbf{26.34} & \textbf{6.60} & 	\textbf{0.9529} & 	0.9804	&  \textbf{3.8151 }& \textbf{29.58}& 	\textbf{4.30 }\\
 
     \hline
     \multirow{7}{*}{PPR10K-b} & UEGAN~\cite{ni2020towards}  & 14.71& 	23.66  & 0.5409	& 0.7695 & 6.9312 & 17.91 & 15.55 \\ 

     & HDRNet~\cite{gharbi2017deep}   &  22.34 & 10.45	& 0.9147 & 0.9579 &	 4.0546  &25.64&	6.78\\
      &  CSRNet~\cite{he2020conditional}  &   23.55 & 9.10& 	0.9347	 &0.9691 & 3.9833  &26.77&	5.91\\

      &  LPTN~\cite{liang2021high}   &  21.86 & 10.02	 &	0.9090	 &0.9440 & 	4.0414 & 25.27&	6.48\\
    &  3D LUT~\cite{zeng2020learning}   &   24.81	&  7.79 &	 0.9458 &	0.9756 &  3.9817  &28.10	&5.04\\ %
      &  3D LUT HRP~\cite{liang2021ppr10k}   &  25.06 & 7.51	&	0.9454 &	\textbf{0.9765}& 	3.9498 & 28.36 &	4.85\\ %
    &  \textbf{Ours} & \textbf{25.33} &\textbf{7.39}& \textbf{0.9482}&	0.9761 &  \textbf{3.8129} & \textbf{28.61} &\textbf{4.79} \\

     \hline
   \multirow{7}{*}{PPR10K-c}  & UEGAN~\cite{ni2020towards}  &15.07 &22.81 & 0.5561  &	0.7756  & 6.9312  & 18.30	 &  14.94\\ 
     &  HDRNet~\cite{gharbi2017deep}   &  23.07	& 9.90& 	0.9112 & 	0.9637 &  3.9471 & 26.40& 	6.44 \\
      &  CSRNet~\cite{he2020conditional}  &  23.33  &9.60 & 0.9162 & 	0.9674 & 	3.9805 &26.68&	6.18\\

      &  LPTN~\cite{liang2021high}   & 21.70 &10.54	& 0.8980 &	0.9406 & 	3.9536 &  25.10&	6.82\\
 &   3D LUT~\cite{zeng2020learning}   &    25.20	& 7.79& 	0.9396 &	0.9758 & 3.9523 & 28.46&	5.05 \\ 
 &  3D LUT HRP~\cite{liang2021ppr10k}   & 25.46	& 7.43& 	0.9388	& \textbf{0.9762} & 3.9217  &  28.79 &	4.82 \\ %
     &  \textbf{Ours} &  \textbf{25.68} & 	\textbf{7.41} &  \textbf{0.9429}&	\textbf{0.9762}&  \textbf{3.7751} & \textbf{28.97}	& \textbf{4.80}  \\
    \bottomrule
  \end{tabular} 
}
\end{table*}

\section{Experiments}
\subsection{Dataset and Evaluation Metrics}
\textbf{PPR10K Dataset. } We employ the 360p (short side) low-resolution version of the PPR10K dataset~\cite{liang2021ppr10k} for training and evaluation.  The PPR10K dataset contains 11,161 photos in total and corresponding references retouched by three experts (a/b/c).  For each photo, a human-region mask ($\boldsymbol{M}$ in Eq.~\ref{mask}) is provided to better use the human-region priority (HRP) strategy.   Following the official setting, we divide 8,875 photos for training and maintain the rest 2,286 photos for testing.
Since the human-region masks provided by the PPR10K dataset depict all human regions rather than instances, it is not suitable for interactive portrait retouching, which requires flexibly retouching different persons.  \red{Therefore, we further prepare human instance masks ($\boldsymbol{M}^{it}$ in Eq.~\ref{inter mask}) with the following steps.  We first apply Cascade Mask R-CNN~\cite{cai2019cascade} to perform object detection.  For the detected multi-person images, Swin Transformer-S~\cite{liu2021swin} is then used to segment the human instances, while the annotations of single-person images are kept for interactive portrait retouching.}

\textbf{Objective Evaluation Metrics.} To evaluate the performance of our automatic retouching branch and compare with state-of-the-art methods, we adopt the following evaluation metrics: PSNR, SSIM~\cite{wang2004image}, MS-SSIM~\cite{wang2003multiscale} and the CIELAB color difference $\Delta E_{a b}$~\cite{liang2021ppr10k}. To evaluate the human-region priority, we follow the previous setting~\cite{liang2021ppr10k} and adopt $PSNR^{HC}$ and $\Delta E_{a b}^{HC}$ with human-centered focus. The general expression of $\Delta E_{a b}^{HC}$ is shown as follows:
\begin{equation}
\Delta E_{a b}^{H C}=\left\|\boldsymbol{W}_{\boldsymbol{I}} \odot \boldsymbol{\tilde{I}^{a b}_y} -\boldsymbol{W}_{\boldsymbol{I}} \odot \boldsymbol{I^{a b}_y} \right\|_{2},
\end{equation}
where $\boldsymbol{\tilde{I}^{a b}_y}$, $\boldsymbol{I^{a b}_y}$ and $\boldsymbol{W}_{\boldsymbol{I}}$ denote the retouched result, the ground truth image in the CIELAB color space and the weighting matrix, respectively.  We set $w_{ij}= 1$ for human-regions, $w_{ij}= 0.5$ for the background, and $\Delta E_{a b}^{H C}$ equals $\Delta E_{a b}$ if $w_{ij}= 1$ is set for all pixels.
Since portrait retouching is a highly harmony-required task, we also include NIQE~\cite{mittal2012making}, a widely-used non-reference image quality assessment metric to evaluate the naturalness of retouched results. Due to the flexibility of interactive portrait retouching, it is not practical to evaluate results with available ground truths in PPR10K dataset, we introduce several non-reference image quality assessment metrics, including NIQE~\cite{mittal2012making}, ILNIQE~\cite{zhang2015feature}, BRISQUE~\cite{mittal2012no}, MA~\cite{ma2017learning}, PI~\cite{blau20182018} and BIQI~\cite{moorthy2009modular} to evaluate the interactive portrait retouching.

\textbf{Subjective Evaluation Metrics.} 
To evaluate the perceptual quality of our interactive retouching results, we perform a user study and invite 20 participants for the subjective evaluation.     For each version of the PPR10K dataset (a/b/c), we randomly select 50 images at a time and make a pairwise comparison between all methods (5 in total), obtaining a total of 1,500 pairwise comparisons for all 150 images.     \red{Each participant is unaware of the portrait retouching task and asked to consider whether the results are visually realistic}, whether there are artifacts, and most importantly, whether the human instance regions are in harmony with each other and the background.     They are required to select the better one from each pair.     Then we record the times that each method is selected.     \red{To make a more intuitive comparison with cascaded baseline methods, we follow the previous methods~\cite{ling2021region,tsai2017deep,guo2021image} to further summarize the user study with the preference rate and exponential Bradley-Terry model~\cite{bradley1952rank} (B-T score) for evaluation.  The exponential Bradley-Terry model is sensitive to selection times, and the B-T score that ranks each baseline grows exponentially over selection times.}

\begin{figure*}[t]\vspace{-6pt}
  \begin{overpic}[width=58pt, clip=true, trim=0 0pt 0  50pt]{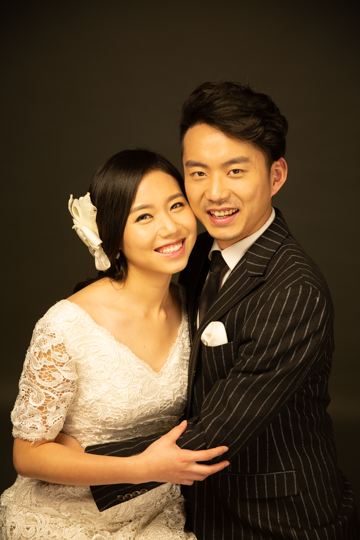}\put(-10,-36){\begin{sideways} \scriptsize{PPR10K-a Dataset}  \end{sideways}}\end{overpic} 
\hspace{-2pt}\includegraphics[width=58pt, clip=true, trim=0 0pt 0  50pt]{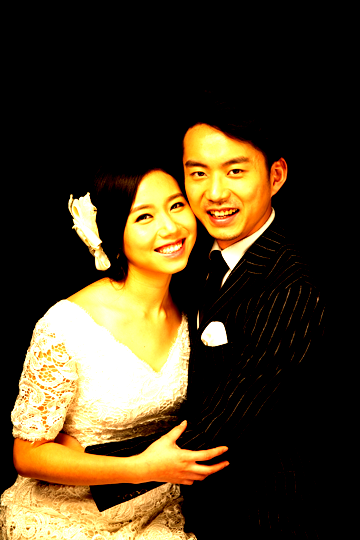}
\includegraphics[width=58pt, clip=true, trim=0 0pt 0  50pt]{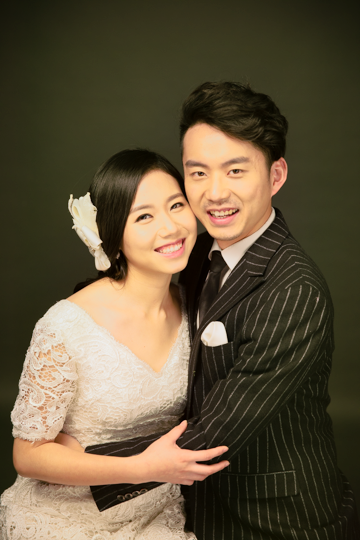}
\includegraphics[width=58pt, clip=true, trim=0 0pt 0  50pt]{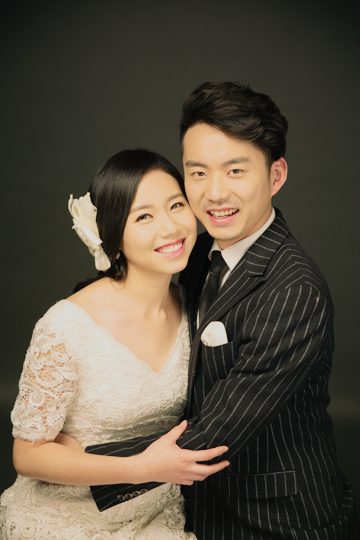}
\includegraphics[width=58pt, clip=true, trim=0 0pt 0  50pt]{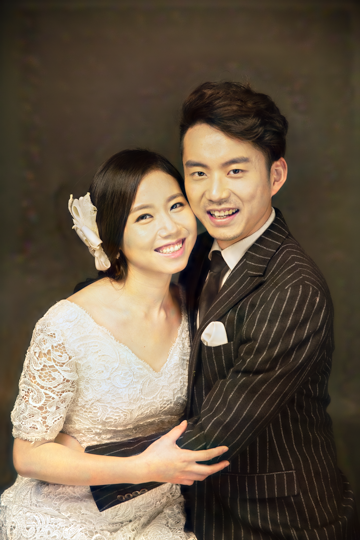}
\includegraphics[width=58pt, clip=true, trim=0 0pt 0  50pt]{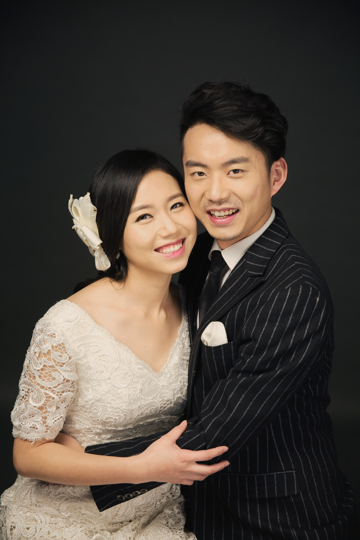}
\includegraphics[width=58pt, clip=true, trim=0 0pt 0  50pt]{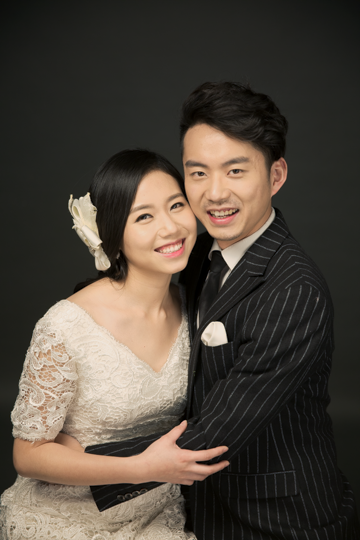}
\includegraphics[width=58pt, clip=true, trim=0 0pt 0  50pt]{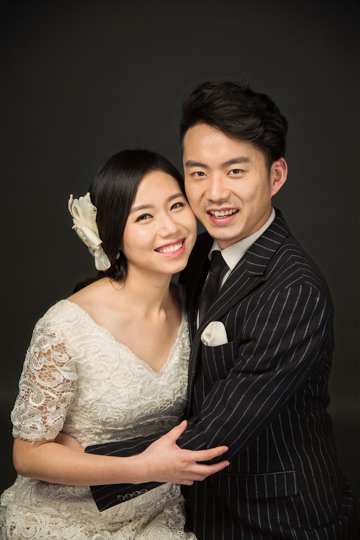}
\includegraphics[width=58pt, clip=true, trim=0 0pt 0  50pt]{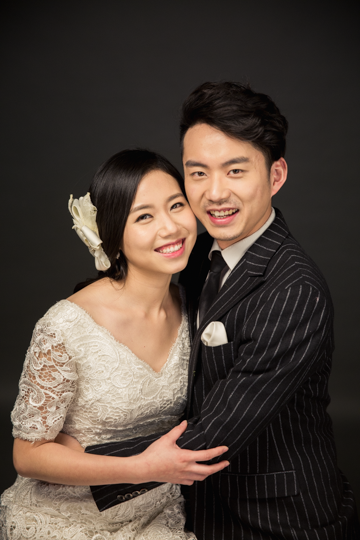}\vspace{2pt}\\
\includegraphics[width=58pt, clip=true, trim=0 0pt 0  60pt]{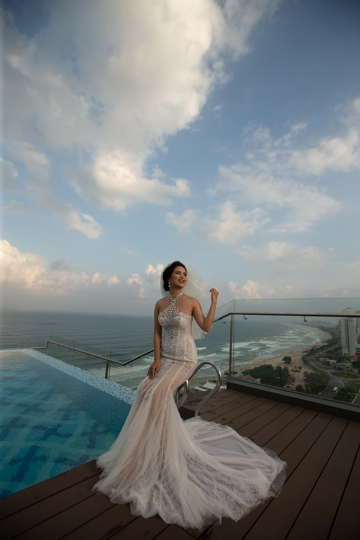}\hspace{-2.5pt}
\includegraphics[width=58pt, clip=true, trim=0 0pt 0  60pt]{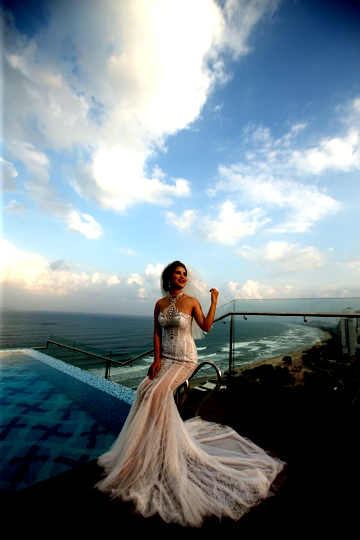}
\includegraphics[width=58pt, clip=true, trim=0 0pt 0  60pt]{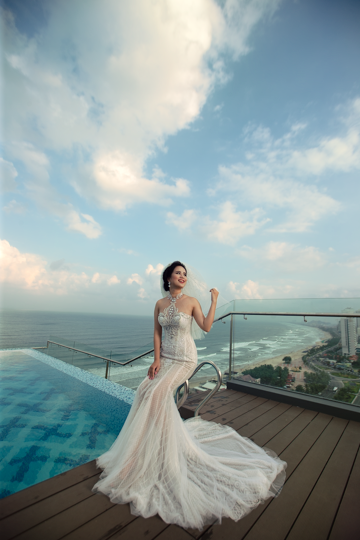}
\includegraphics[width=58pt, clip=true, trim=0 0pt 0  60pt]{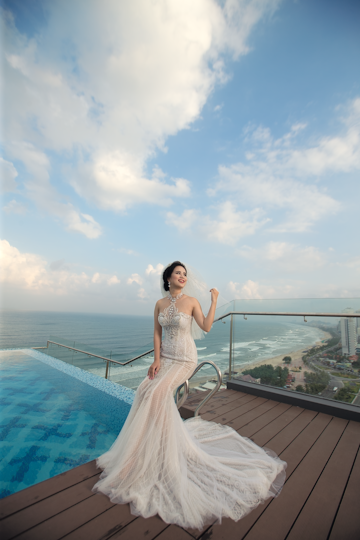}
\includegraphics[width=58pt, clip=true, trim=0 0pt 0  60pt]{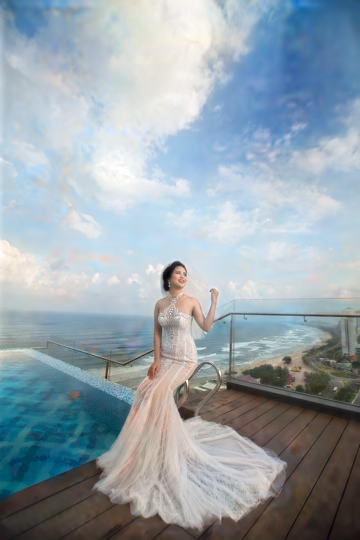}
\includegraphics[width=58pt, clip=true, trim=0 0pt 0  60pt]{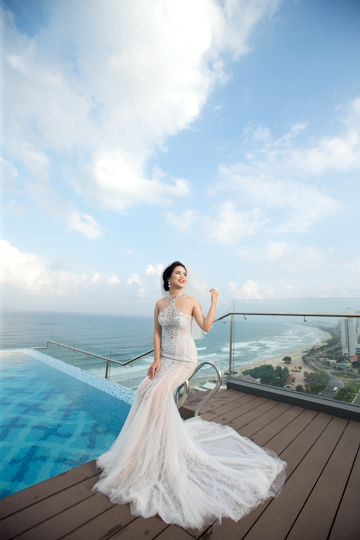}
\includegraphics[width=58pt, clip=true, trim=0 0pt 0  60pt]{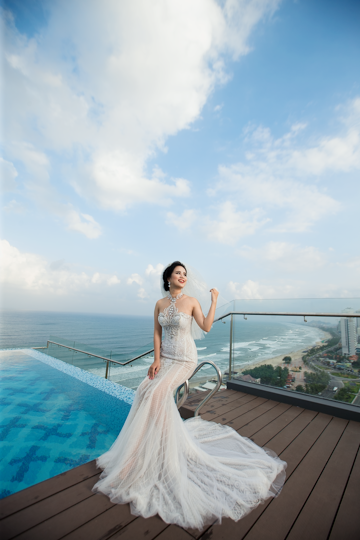}
\includegraphics[width=58pt, clip=true, trim=0 0pt 0  60pt]{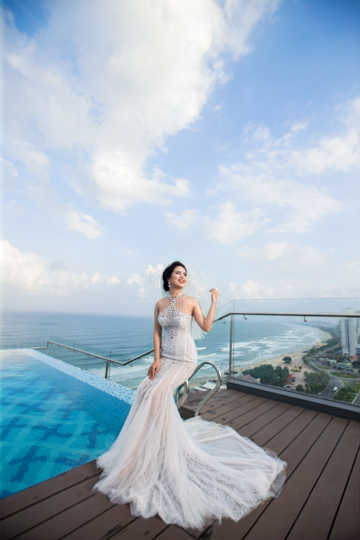}
\includegraphics[width=58pt, clip=true, trim=0 0pt 0  60pt]{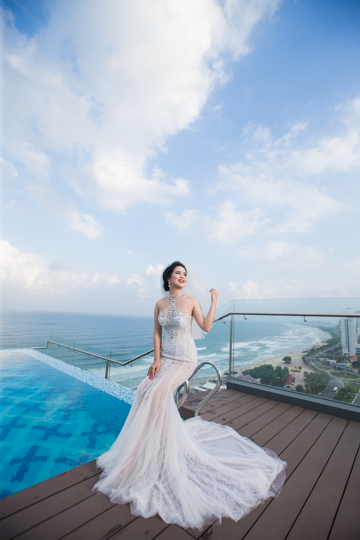}

\vspace{2pt}\begin{overpic}[width=58pt, clip=true, trim=0 50pt 0  20pt]{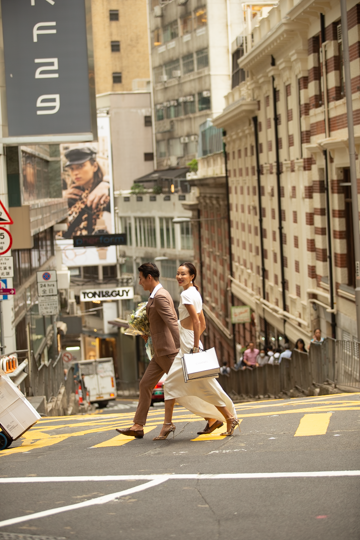}\put(-10,-36){\begin{sideways} \scriptsize{PPR10K-b Dataset}  \end{sideways}}\end{overpic} 
\hspace{-2pt}\includegraphics[width=58pt, clip=true, trim=0 50pt 0  20pt]{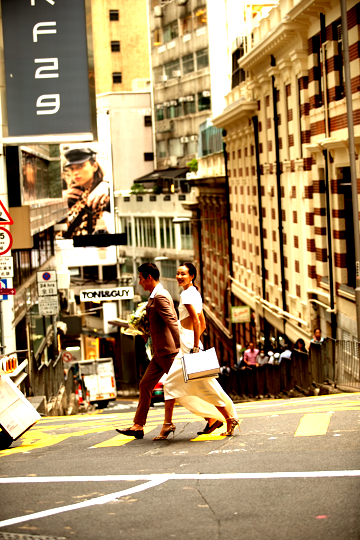}
\includegraphics[width=58pt, clip=true, trim=0 50pt 0  20pt]{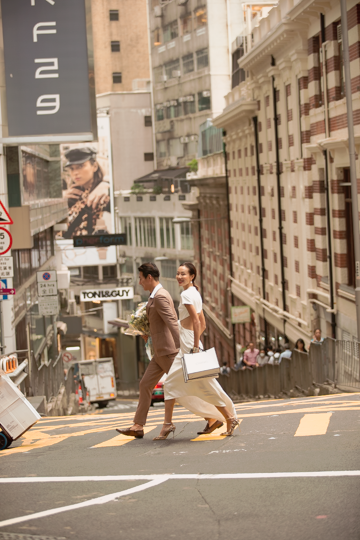}
\includegraphics[width=58pt, clip=true, trim=0 50pt 0  20pt]{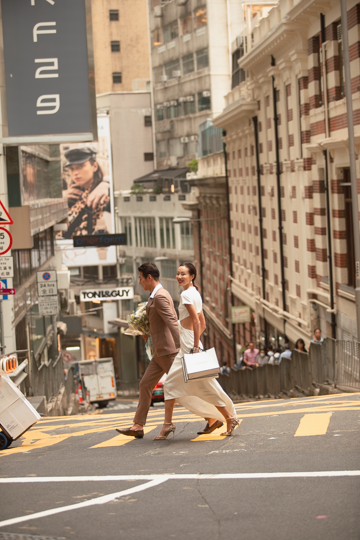}
\includegraphics[width=58pt, clip=true, trim=0 50pt 0  20pt]{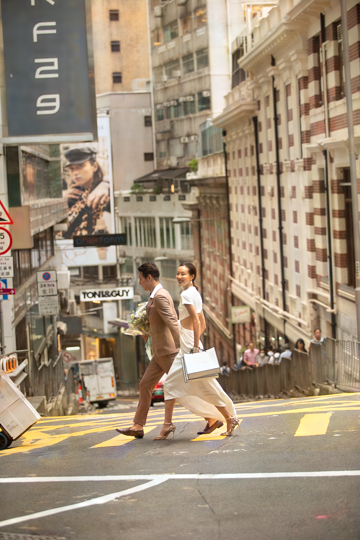}
\includegraphics[width=58pt, clip=true, trim=0 50pt 0  20pt]{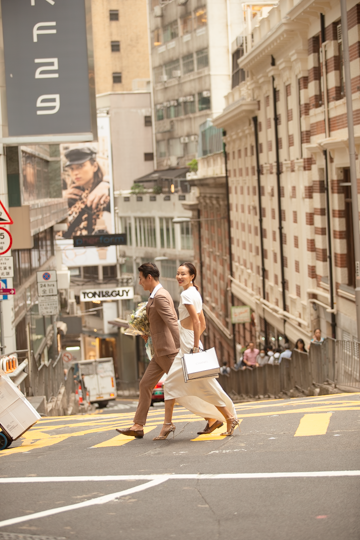}
\includegraphics[width=58pt, clip=true, trim=0 50pt 0  20pt]{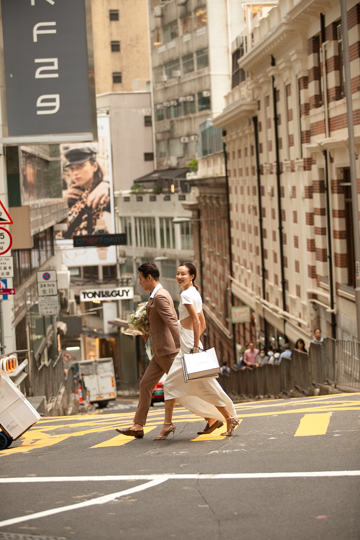}
\includegraphics[width=58pt, clip=true, trim=0 50pt 0  20pt]{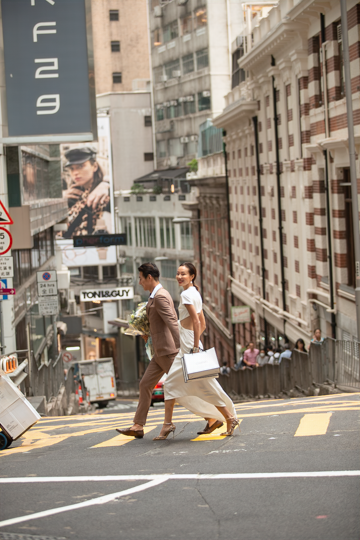}
\includegraphics[width=58pt, clip=true, trim=0 50pt 0  20pt]{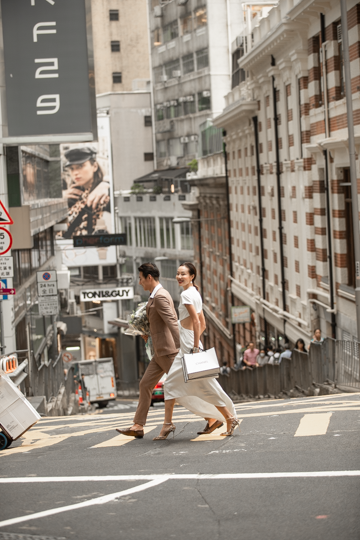}\vspace{2pt}\\
\includegraphics[width=58pt, clip=true, trim=150pt 0 110pt  0]{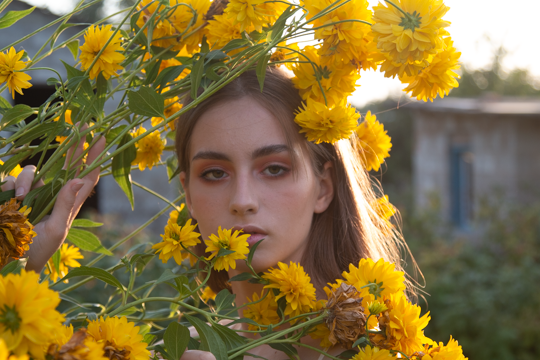}\hspace{-2.5pt}
\includegraphics[width=58pt, clip=true, trim=150pt 0 110pt  0]{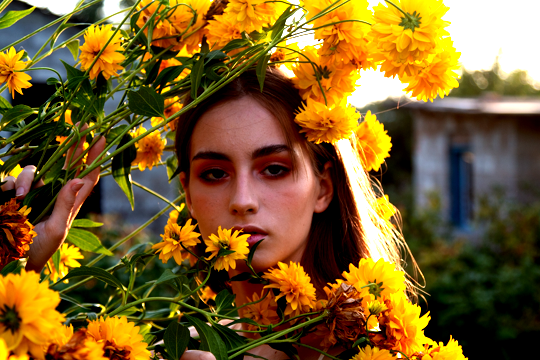}
\includegraphics[width=58pt, clip=true, trim=150pt 0 110pt  0]{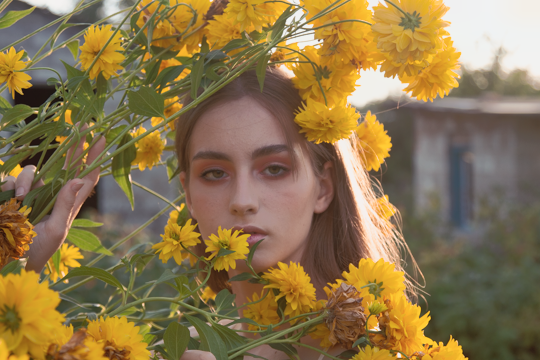}
\includegraphics[width=58pt, clip=true, trim=150pt 0 110pt  0]{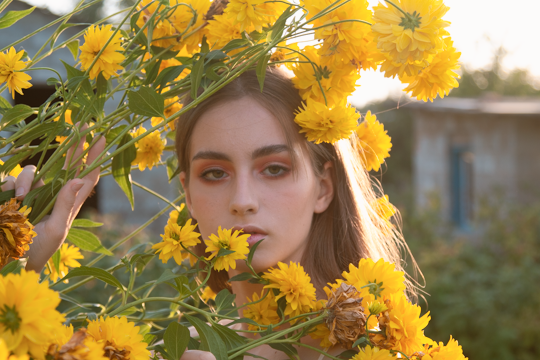}
\includegraphics[width=58pt, clip=true, trim=150pt 0 110pt  0]{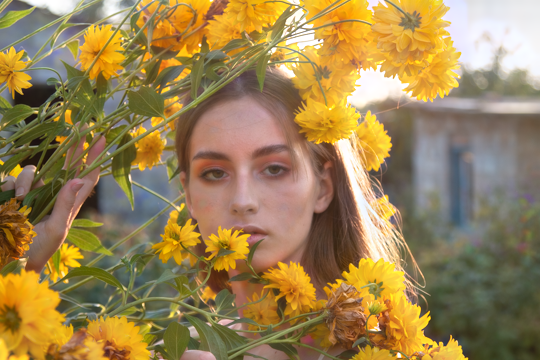}
\includegraphics[width=58pt, clip=true, trim=150pt 0 110pt  0]{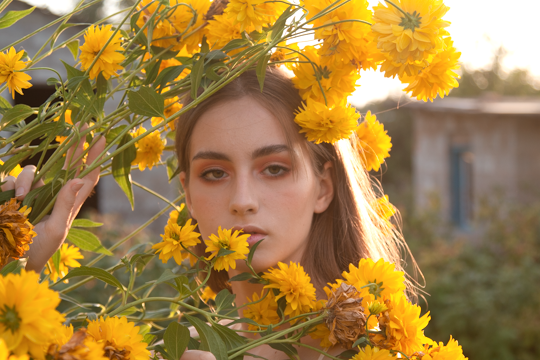}
\includegraphics[width=58pt, clip=true, trim=150pt 0 110pt  0]{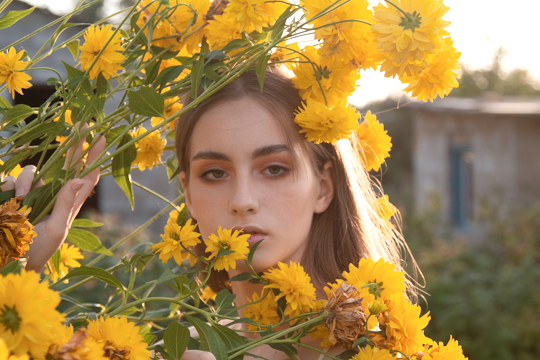}
\includegraphics[width=58pt, clip=true, trim=150pt 0 110pt  0]{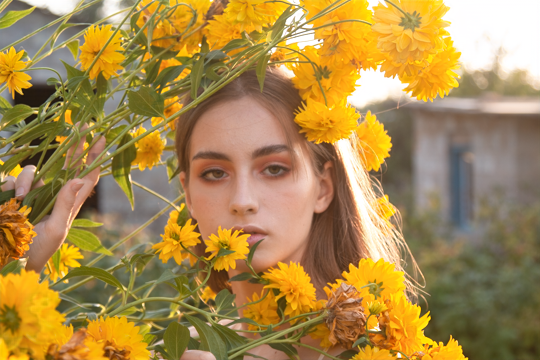}
\includegraphics[width=58pt, clip=true, trim=150pt 0 110pt  0]{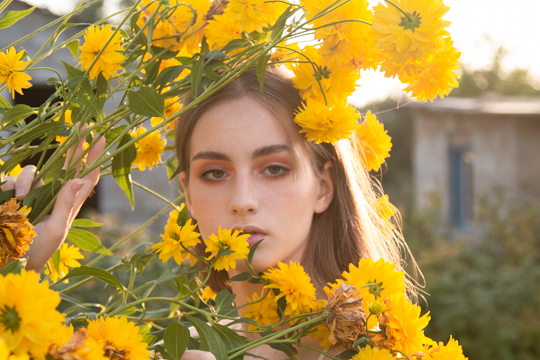}

\vspace{2pt}\begin{overpic}[width=58pt, clip=true, trim=0 100pt 0  0]{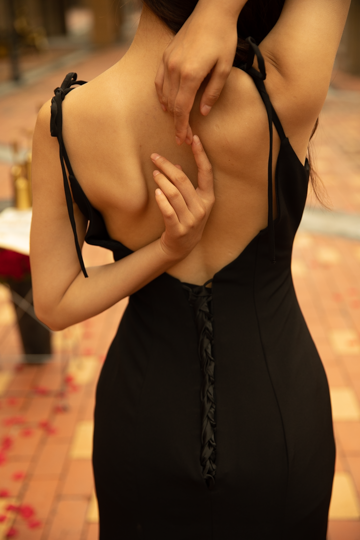}\put(-10,-36){\begin{sideways} \scriptsize{PPR10K-c Dataset}  \end{sideways}}\end{overpic} 
\hspace{-2pt}\includegraphics[width=58pt, clip=true, trim=0  100pt 0  0]{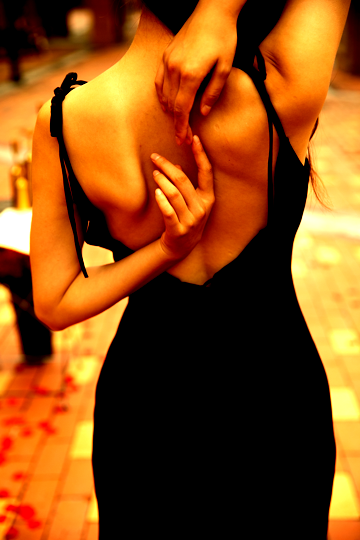}
\includegraphics[width=58pt, clip=true, trim=0 100pt 0  0]{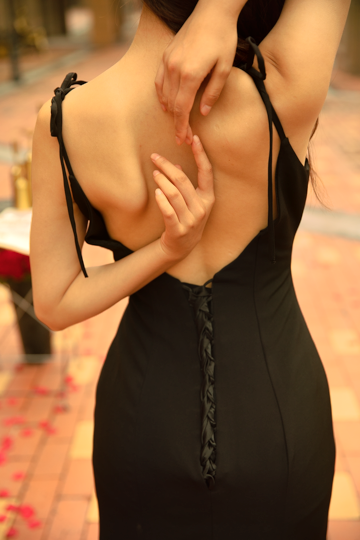}
\includegraphics[width=58pt, clip=true, trim=0 100pt 0  0]{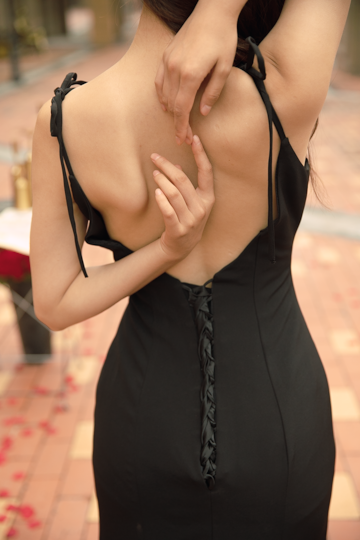}
\includegraphics[width=58pt, clip=true, trim=0 100pt 0  0]{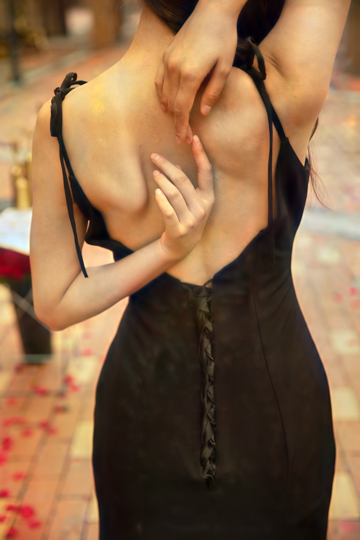}
\includegraphics[width=58pt, clip=true, trim=0 100pt 0  0]{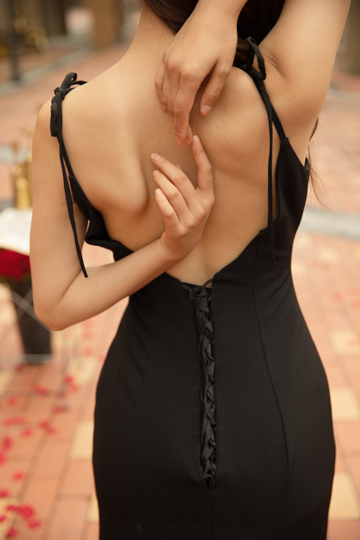}
\includegraphics[width=58pt, clip=true, trim=0 100pt 0  0]{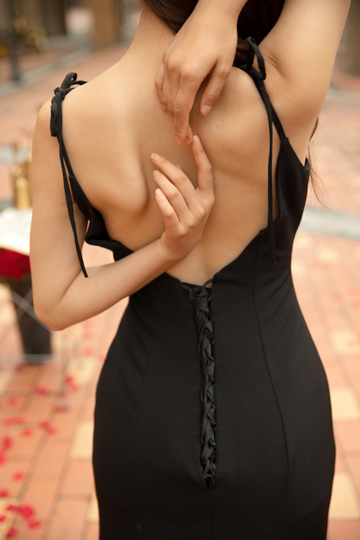}
\includegraphics[width=58pt, clip=true, trim=0 100pt 0  0]{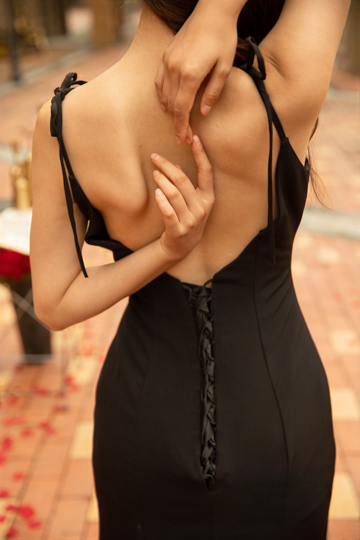}
\includegraphics[width=58pt, clip=true, trim=0 100pt 0  0]{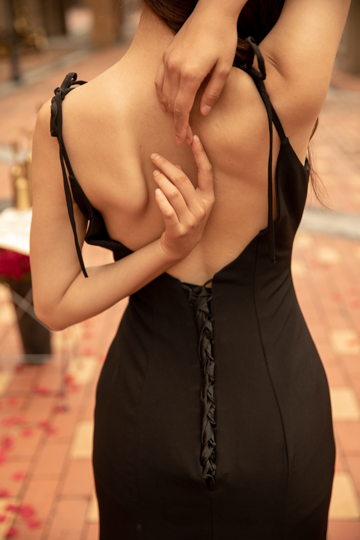}\vspace{2pt}\\
\includegraphics[width=58pt, clip=true, trim=150pt 0 110pt  0]{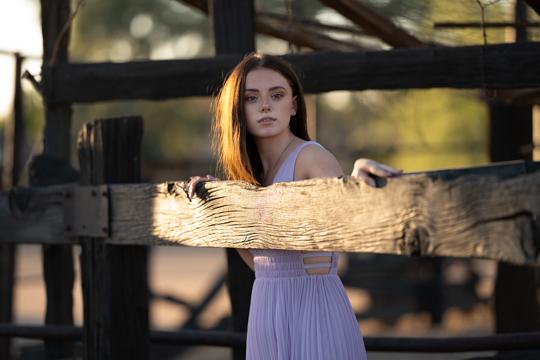}\hspace{-2.5pt}
\includegraphics[width=58pt, clip=true, trim=150pt 0 110pt  0]{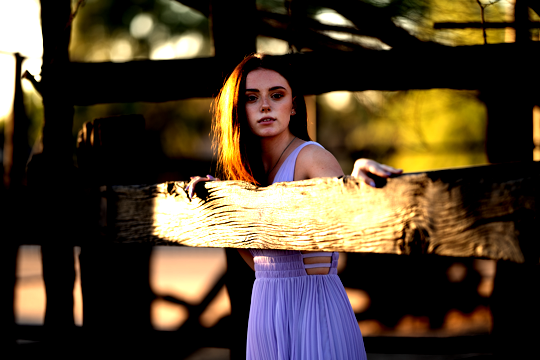}
\includegraphics[width=58pt, clip=true, trim=150pt 0 110pt  0]{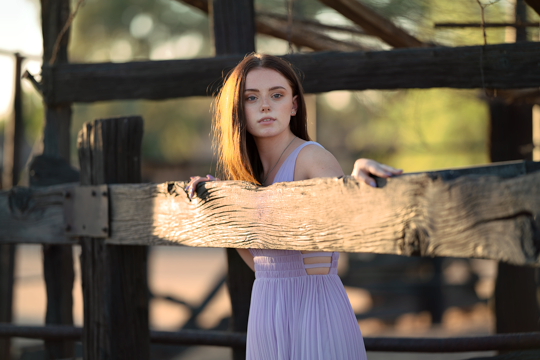}
\includegraphics[width=58pt, clip=true, trim=150pt 0 110pt  0]{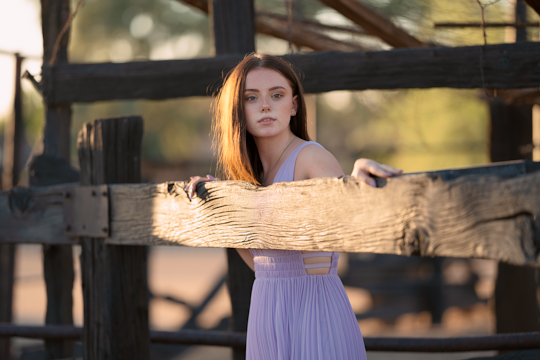}
\includegraphics[width=58pt, clip=true, trim=150pt 0 110pt  0]{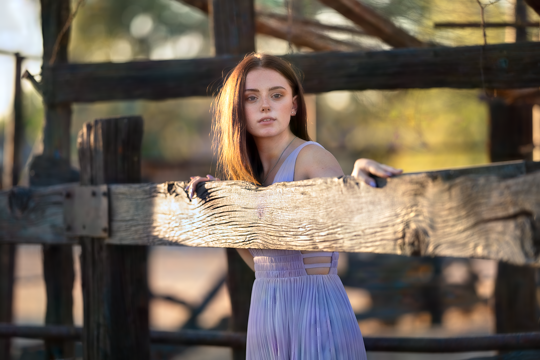}
\includegraphics[width=58pt, clip=true, trim=150pt 0 110pt  0]{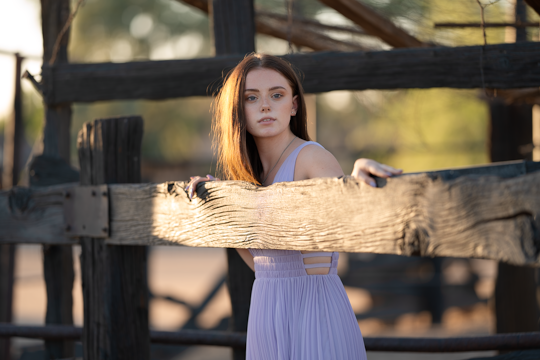}
\includegraphics[width=58pt, clip=true, trim=150pt 0 110pt  0]{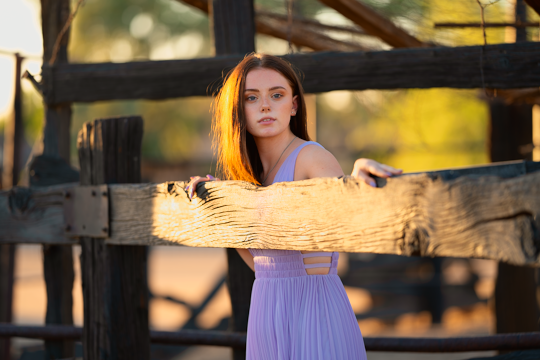}
\includegraphics[width=58pt, clip=true, trim=150pt 0 110pt  0]{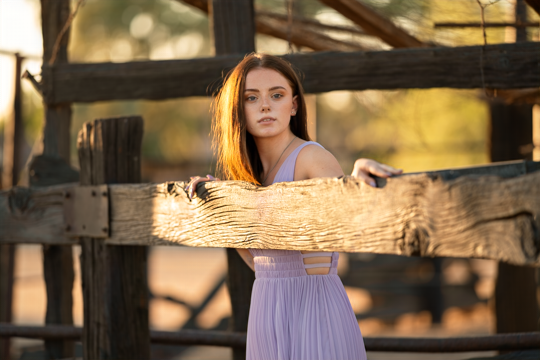}
\includegraphics[width=58pt, clip=true, trim=150pt 0 110pt  0]{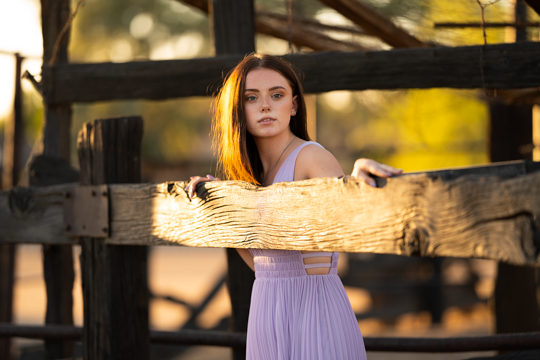}

\hspace{10pt} \scriptsize{(a)~Input}
\hspace{24pt} \scriptsize{(b)~UEGAN}
\hspace{18pt} \scriptsize{(c)~HDRNet}
\hspace{18pt} \scriptsize{(d)~CSRNet }
\hspace{22pt} \scriptsize{(e)~LPTN}
\hspace{24pt} \scriptsize{(f)~3D LUT}
\hspace{9pt} \scriptsize{(g)~3D LUT HRP}
\hspace{16pt} \scriptsize{(h)~Ours}
\hspace{30pt} \scriptsize{(i)~GT}

\vspace{-4pt}\caption{Visual comparison of automatic portrait retouching on the PPR10K-a dataset, PPR10K-b dataset and PPR10K-c dataset.}\vspace{-10pt}
\label{figure: Visualization Comparison}
\end{figure*}

\begin{table*}[t] 
    \centering
    \vspace{-6pt}\caption{Quantitative comparison with cascaded baseline methods on PPR10K dataset. Multiple non-reference image quality assessments are included for comprehensive illustration. \red{We summarize the user study of Fig.~\ref{fig:user study} with preference rate and B-T score.} } 
    \label{table: user study}
     \setlength{\tabcolsep}{2.2mm}{
    \begin{tabular}{c|c|c|c|c|c|c|c|>{\columncolor{lightgrey!40}}c|>{\columncolor{lightgrey!40}}c}
    \toprule
    Dataset & Method    & NIQE $\downarrow$ &  ILNIQE $\downarrow$ & BRISQUE $\downarrow$& MA $\uparrow$ & PI $\downarrow$  & BIQI $\downarrow$ &\red{Preference rate $\uparrow$} & B-T score $\uparrow$\\
        \hline
        \multirow{5}{*}{PPR10K-a} & HDRNet~\cite{gharbi2017deep}  & 4.4818  &  29.5940 &   28.5967&  7.3338&   3.5740& 33.3436 & \red{4.80\%} &  0.369  \\
       &  CSRNet~\cite{he2020conditional}  &  4.4958  & 29.7355 & 28.7410 & 7.3085 & 3.5937  & 33.3779  & \red{12.20\%}  & 0.943\\
       &  3D LUT~\cite{zeng2020learning} & 4.4831  & 29.6556 &  28.6066 &  7.3149 &   3.5841  & 33.3726 &  \red{18.40\%} & 1.934 \\
        & 3D LUT HRP~\cite{liang2021ppr10k}   & 4.4740 & 29.5943  & 28.5860 & 7.3246 &   3.5747 & 33.3403 & \red{28.00\%} & 6.086\\
        & Ours  & \textbf{3.8105} &  \textbf{24.4520} & \textbf{19.2800} & \textbf{8.2226}& \textbf{2.7940} & \textbf{27.8261}& \red{\textbf{36.60\%}}  & \textbf{20.253}    \\
        \hline
         \multirow{5}{*}{PPR10K-b} & HDRNet~\cite{gharbi2017deep}  & 4.4868 & 29.5590 &     28.6078  & 7.3309 & 3.5780 & 33.3392 &  \red{4.00\%} & 0.367 \\
       &  CSRNet~\cite{he2020conditional}   & 4.4841 & 29.6234 &  28.5537  & 7.3110 &  3.5866 & 33.3752 & \red{11.80\%} &	1.046 \\
       &  3D LUT~\cite{zeng2020learning}  & 4.4779 &29.6233 &   28.5175 & 7.3104 & 3.5838 & 33.3804  &  \red{21.20\%} &  3.156\\
        & 3D LUT HRP~\cite{liang2021ppr10k} & 4.4699 & 29.5547 &  28.4554 & 7.3256 &     3.5722 &  33.3446 &  \red{27.00\%} & 	6.198  \\
        & Ours & \textbf{3.8067} & \textbf{24.4442} &  \textbf{19.8518}  & \textbf{8.2234} &     \textbf{2.7917} & \textbf{27.3495 }  &  \red{\textbf{36.00\%}}  & \textbf{20.103} \\
        \hline
         \multirow{5}{*}{PPR10K-c} & HDRNet~\cite{gharbi2017deep}  & 5.2017 & 29.5964 & 28.4960 &7.3238  & 3.9390 & 33.3731  & \red{4.00\%} & 0.372\\
       &  CSRNet~\cite{he2020conditional} & 4.4848 & 29.7199 &  28.5776 &  7.3041 & 3.5904 & 33.3898& \red{11.60\%} & 1.044  \\
       &  3D LUT~\cite{zeng2020learning}& 4.4836 & 29.7277 &28.5342  & 7.2988 &   3.5924& 33.4226  & \red{20.20\%}  & 	2.939\\
        & 3D LUT HRP~\cite{liang2021ppr10k}  & 5.2016 & 29.6558 &  28.5119 & 7.3105 & 3.9456 & 33.3880 & \red{28.80\%} & 8.200	\\
        & Ours	 & \textbf{3.7768} & \textbf{24.9507} & \textbf{19.4812} & \textbf{8.2514} & \textbf{2.7627} & \textbf{28.2272} & \red{\textbf{35.40\%}} & \textbf{19.595}  \\
    \bottomrule
    \end{tabular} 
    } 
\end{table*}

\subsection{Baseline Methods and Implementation Details}
\textbf{Automatic Retouching Baselines.}
We select the following representative photo retouching methods for comparison on the automatic retouching task: 
UEGAN~\cite{ni2020towards}, HDRNet~\cite{gharbi2017deep}, CSRNet~\cite{he2020conditional}, LPTN~\cite{liang2021high}, 3D LUT~\cite{zeng2020learning}, and 3D LUT HRP~\cite{liang2021ppr10k}, Among these baseline methods, 3D LUT HRP adopts the human-region priority strategy to achieve region-awareness, while the others do not have special designs for the portrait retouching task.

\textbf{Interactive Retouching Baselines.}
Since there are few learning-based interactive retouching methods, to evaluate the performance of our interactive retouching branch, we design a cascaded strategy of \textit{``interactive segmentation - local retouching - image harmonization"} (demonstrated in Fig.~\ref{fig:teaser}(b)).  \red{ To be consistent with our interactive setting, the interactive segmentation network also takes clicks as user guidance and outputs an accurate mask of the user-specified region.  The locally retouched result is obtained by pasting the automatically retouched result of the user-specified region to the raw input. The harmonization network adjusts the locally retouched result to achieve naturalness.} \red{We utilize RITM-H18~\cite{sofiiuk2021reviving} as the interactive segmentation model, which is trained on a segmentation-specific dataset~\cite{sofiiuk2021reviving}.  HDRNet~\cite{gharbi2017deep}, CSRNet~\cite{he2020conditional}, 3D LUT~\cite{zeng2020learning} and 3D LUT HRP~\cite{liang2021ppr10k} are chosen to perform retouching.  DHT~\cite{guo2021image} is employed as the harmonization network.}

\textbf{Implementation Details.}
\orange{To simulate the user interaction in interactive retouching, we follow RITM~\cite{sofiiuk2021reviving} to randomly place positive/negative clicks in/out of the human instance region.   For each type of click, the number of clicks is randomly generated between 0 and 5 (the generated clicks may contain only one kind).} \purple{To convert coordinate-based positive/negative clicks into binary user guidance map $G_p$/$G_n$}, \green{in the area centered at positive/negative clicks and with a radius of 3, the pixels are assigned 1, and the others are assigned 0.}  During training, we set the batch size to 10 and adopt the Adam optimizer. The learning rate is initialized as $10^{-4}$, and decays by 0.5 for every $10^{5}$ iterations. The whole training process takes $4 \times 10^{5}$ iterations, of which the automatic retouching training, the interactive retouching training and the joint training of two branches take up $3 \times 10^{5}$, $4 \times 10^{4}$ and $6 \times 10^{4}$ iterations, respectively. Our model is implemented with the PyTorch framework and all experiments are performed on a single NVIDIA GTX2080Ti GPU.

\subsection{Comparison with State-of-the-art Methods}
\subsubsection{Automatic Region-aware Retouching} 
\
\par \textbf{Quantitative Comparison.}
The main results are listed in Table~\ref{quantitative conmparison}, where we mark the best results in bold. We can observe that the proposed method achieves the best performance in terms of most quantitative metrics. UEGAN learns the mapping with unpaired learning and fails to obtain a proper mapping function, thus, performing the worst. LPTN neglects the high fidelity requirement of the retouching task, resulting in  poor performance. As 3D LUT HRP applies an additional human-region priority strategy on 3D LUT and forces more attention on human regions, it has better quantitative performance than 3D LUT. HDRNet and CSRNet show comparable performance. In contrast, our method properly learns global retouching and pays special attention to human-regions, referring to the human-centered assessments of $PSNR^{HC}$ and $\Delta E_{a b}^{HC}$.

\textbf{Visual Comparison.}
The visual results on the PPR10K dataset are shown in Fig.~\ref{figure: Visualization Comparison}. UEGAN inappropriately learns the mapping function and introduces overexposure problem. HDRNet tends to squeeze the dynamic range to a middle level. Since CSRNet tones inputs globally, it significantly promotes the global brightness but fails to keep a balance between the portraits and the background. \red{Because inputs are decomposed into components of low and high frequency, the results of LPTN contain unsatisfying artifacts} (e.g., artifact on the girl's dress in the last example). 
Based on 3D LUT, 3D LUT HRP achieves particular human-region retouching with the human-region priority strategy. Under the condition of completing subtasks with a unified framework, the results of our proposed method are more natural and visually closer to the ground truth than 3D LUT HRP.

\subsubsection{Interactive Region-aware Retouching}
\
\par
\begin{figure}[t]
    \centering
    \includegraphics[width=250pt, clip=true, trim=5pt 12pt  5pt 0]{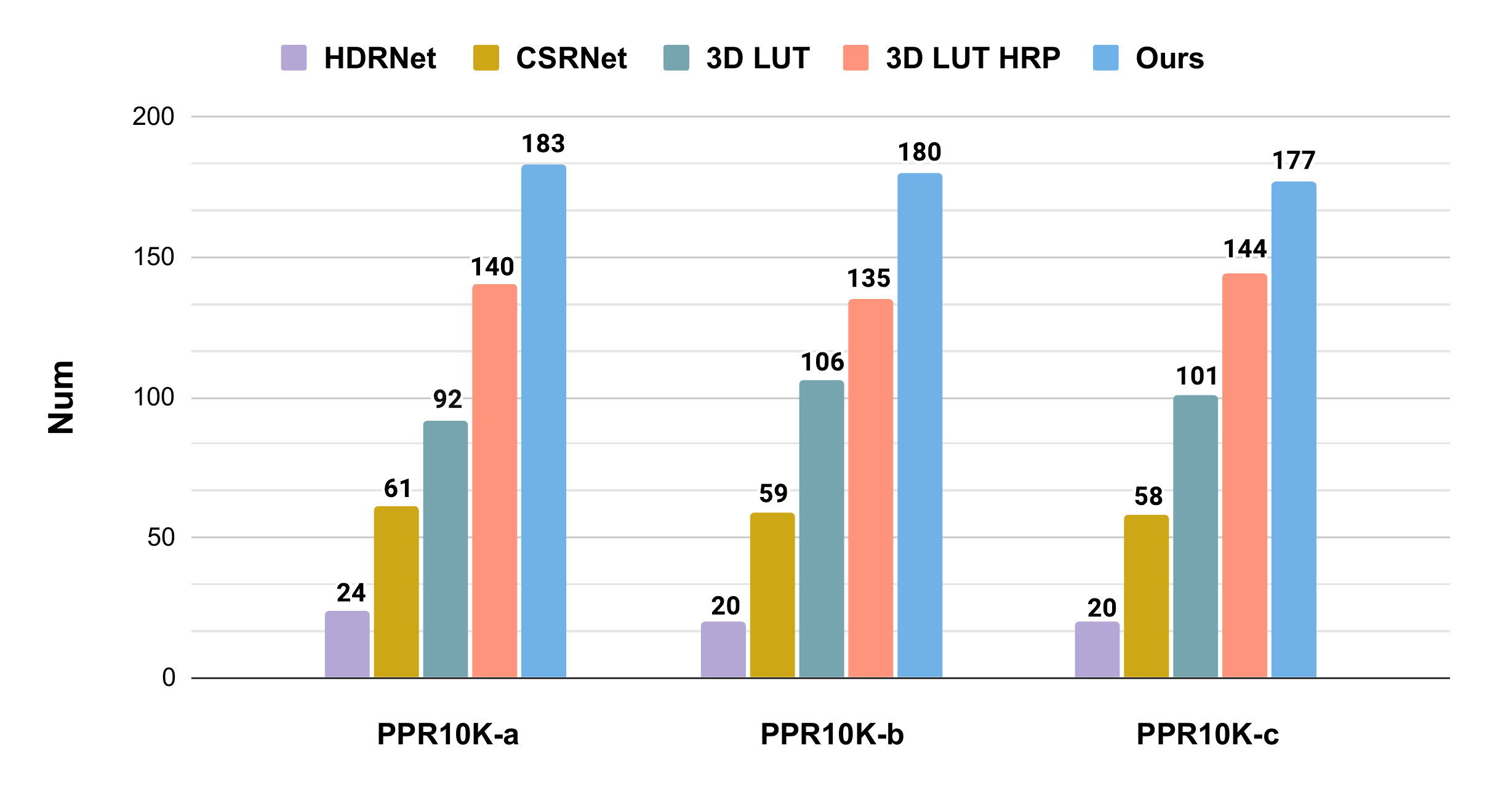}\vspace{-12pt}
    \caption{Results of user study. For simplicity, we distinguish the cascaded retouching baselines from each other with adopted retouching methods, while keeping RITM-H18~\cite{sofiiuk2021reviving} and DHT~\cite{guo2021image} as the interactive segmentation and image harmonization by default. The number of the user selections is displayed at the top of each method. Our method achieves the most favorable results.} 
    \label{fig:user study}
\end{figure}

 \textbf{User Study.}
We conduct a user study to evaluate the perceptual quality of interactive region-aware retouching results.   The cascaded interactive retouching baselines are distinguished by the adopted retouching methods (e.g., HDRNet, CSRNet).
The times that each method is selected during the pairwise comparison are reported in Fig.~\ref{fig:user study}, showing that our method obtains the most user selections.    \red{The preference rate that reflects the proportions of selected times and compared times for each method is included in Table~\ref{table: user study}.   As can be seen, compared to the cascaded interactive retouching baselines, our results are preferred by most of the users on three subsets.   The B-T score that reports the rank of each method is reported in Table~\ref{table: user study}, which shows that our method ranks the highest among baseline methods.}

\begin{figure*}[t]
  \centering
\vspace{-6pt}\begin{overpic}[width=85pt]{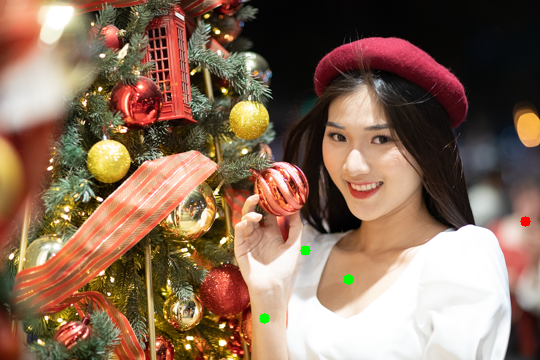}\put(-10,-65){\begin{sideways} \scriptsize{PPR10K-a Dataset}  \end{sideways}}\end{overpic} 
\hspace{-7pt} \includegraphics[width=85pt]{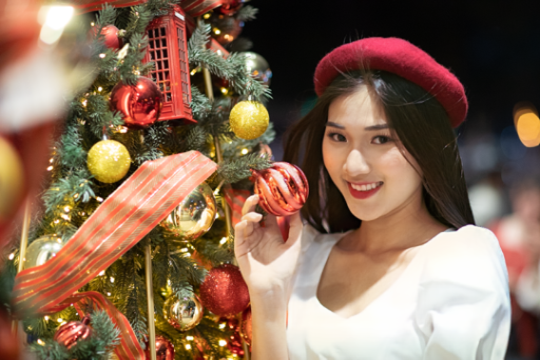}
\includegraphics[width=85pt]{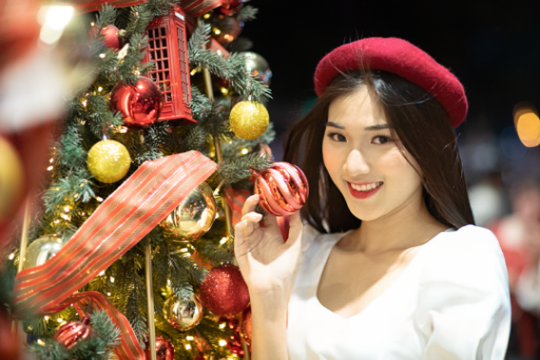}
\includegraphics[width=85pt]{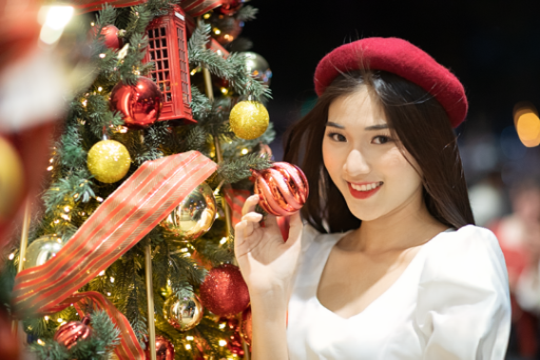}
\includegraphics[width=85pt]{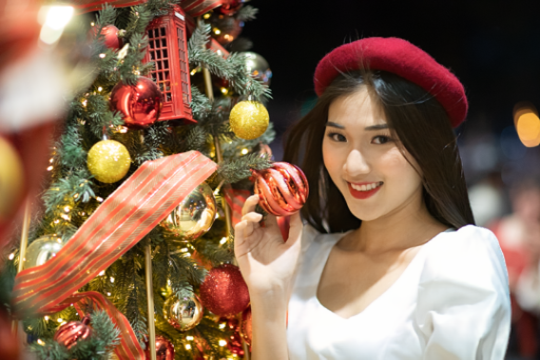}
\includegraphics[width=85pt]{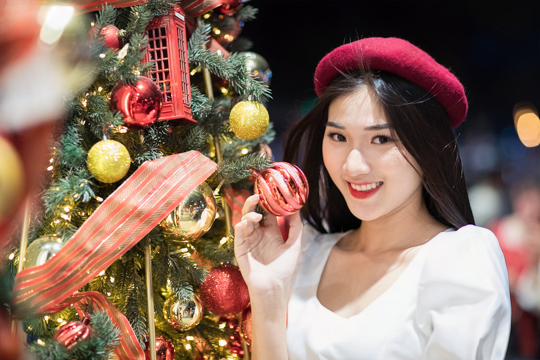} \\

\hspace{1pt}\scriptsize{2.9656} 
\hspace{63pt}\scriptsize{3.4665} 
\hspace{63pt}\scriptsize{3.5757} 
\hspace{63pt}\scriptsize{3.5589} 
\hspace{63pt}\scriptsize{3.5443} 
\hspace{63pt}\scriptsize{2.8074}\vspace{1pt}  

\includegraphics[width=85pt]{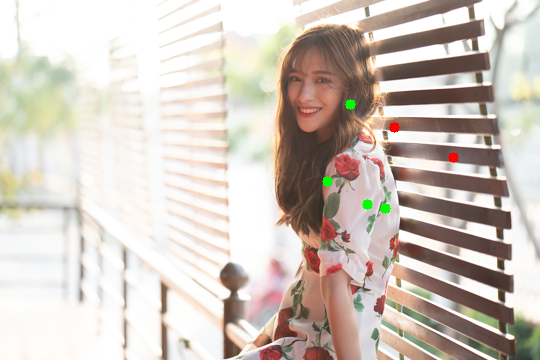}
\includegraphics[width=85pt]{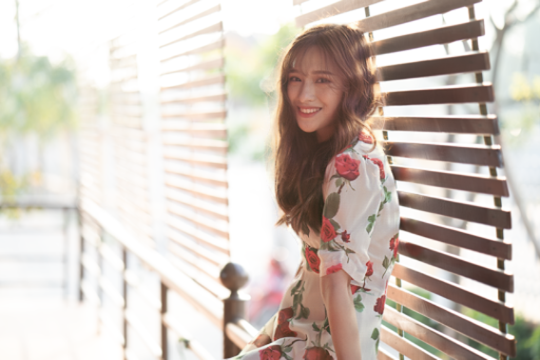}
\includegraphics[width=85pt]{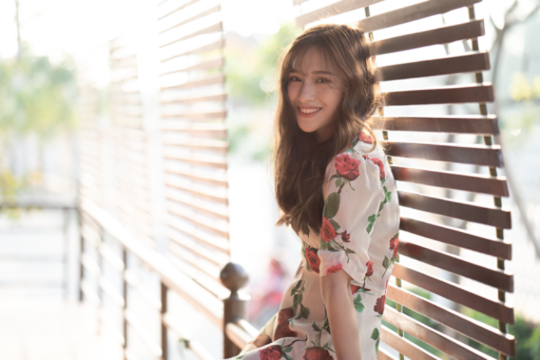}
\includegraphics[width=85pt]{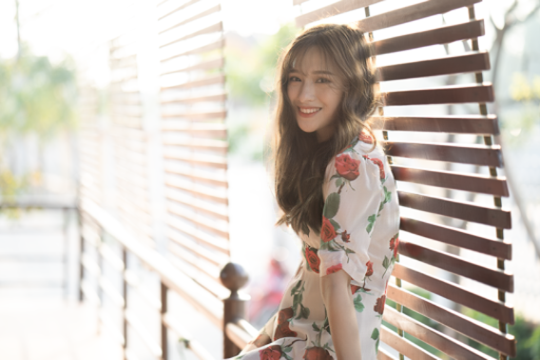}
\includegraphics[width=85pt]{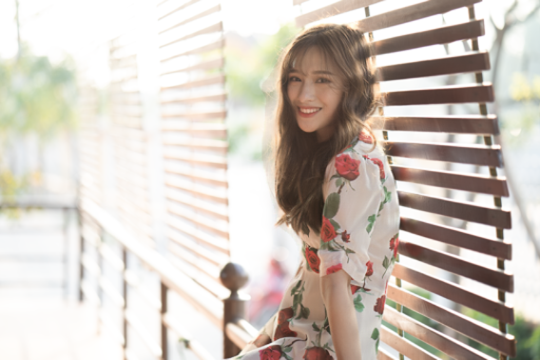}
\includegraphics[width=85pt]{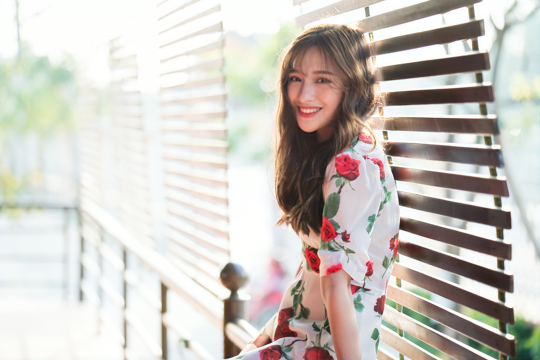} \\

\hspace{1pt}\scriptsize{4.1514}
\hspace{63pt}\scriptsize{4.1234}
\hspace{63pt}\scriptsize{4.2290}
\hspace{63pt}\scriptsize{4.2643}
\hspace{63pt}\scriptsize{4.2384}
\hspace{63pt}\scriptsize{3.9225}\vspace{1pt}

\begin{overpic}[width=85pt]{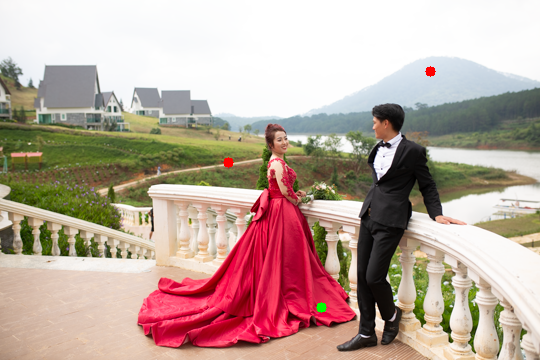}\put(-10,-32){\begin{sideways} \scriptsize{PPR10K-b Dataset}  \end{sideways}}\end{overpic} 
\hspace{-5pt} \includegraphics[width=85pt]{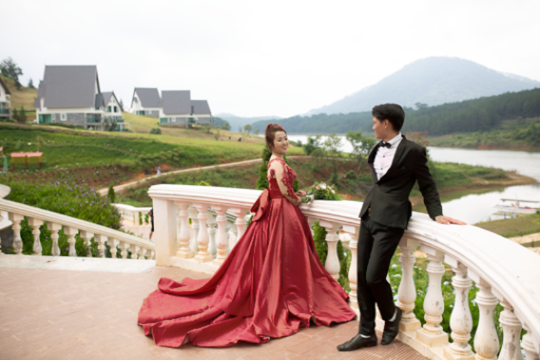}
\includegraphics[width=85pt]{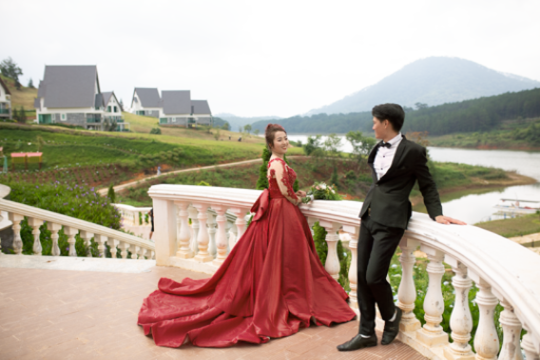}
\includegraphics[width=85pt]{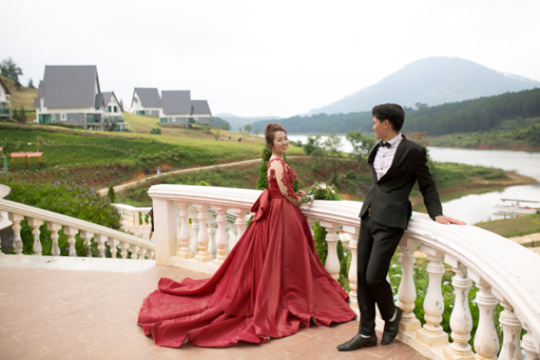}
\includegraphics[width=85pt]{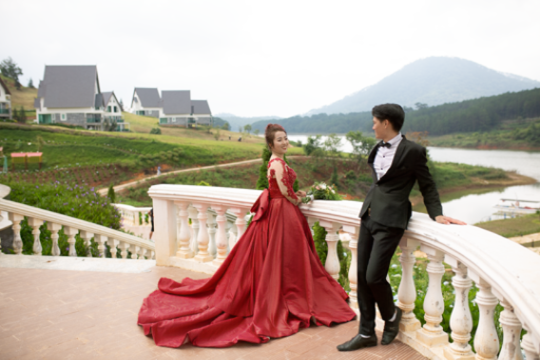}
\includegraphics[width=85pt]{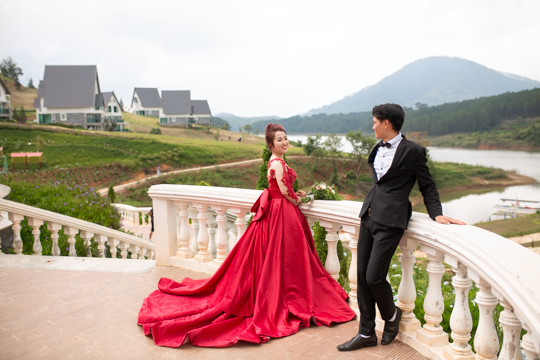} \\

\hspace{1pt}\scriptsize{2.7730}
\hspace{63pt}\scriptsize{3.4785}
\hspace{63pt}\scriptsize{3.5232}
\hspace{63pt}\scriptsize{3.5169}
\hspace{63pt}\scriptsize{3.4798}
\hspace{63pt}\scriptsize{2.6690}\vspace{1pt}

\includegraphics[width=85pt]{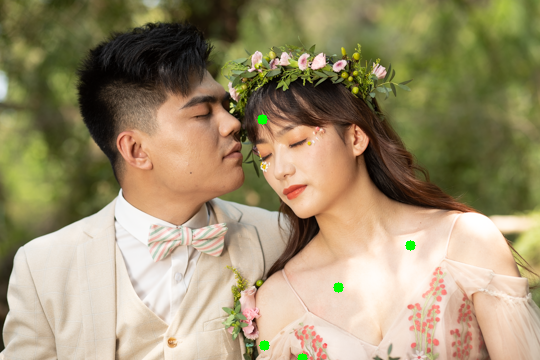}
\includegraphics[width=85pt]{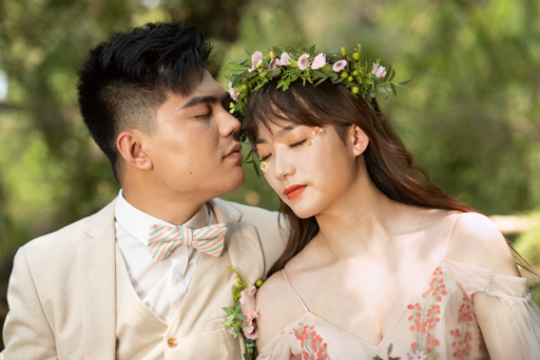}
\includegraphics[width=85pt]{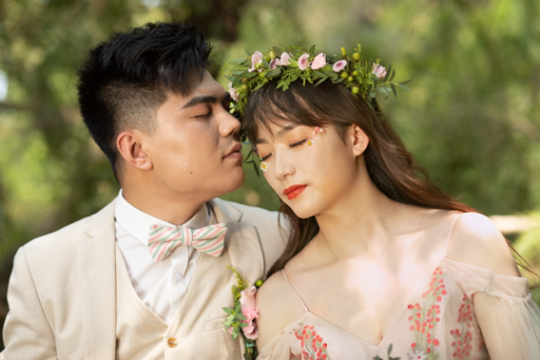}
\includegraphics[width=85pt]{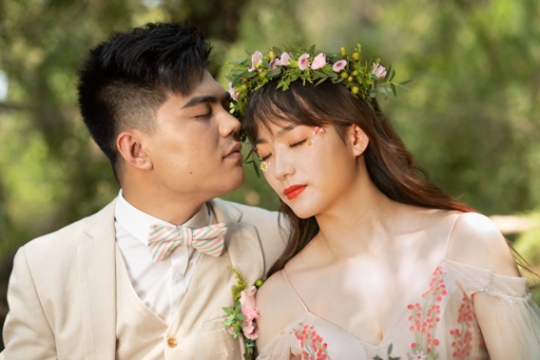}
\includegraphics[width=85pt]{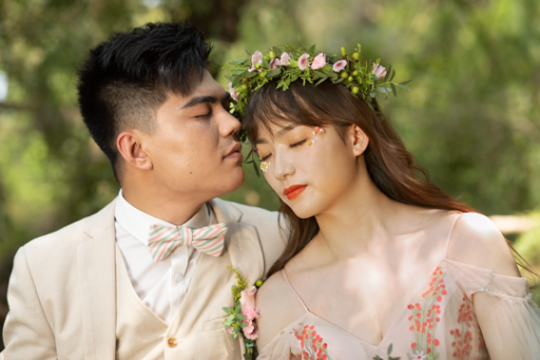}
\includegraphics[width=85pt]{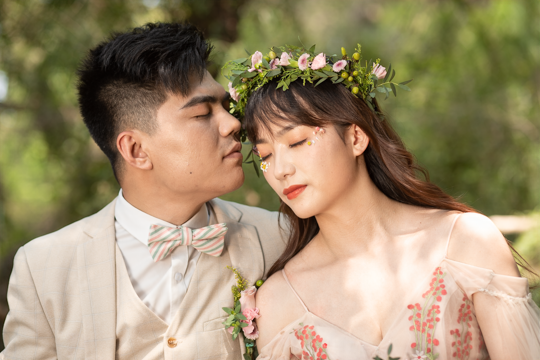} 

\hspace{1pt}\scriptsize{4.0024}
\hspace{63pt}\scriptsize{4.6935}
\hspace{63pt}\scriptsize{4.6895}
\hspace{63pt}\scriptsize{4.6651}
\hspace{63pt}\scriptsize{4.7109}
\hspace{63pt}\scriptsize{3.8272}\vspace{1pt}

\begin{overpic}[width=85pt]{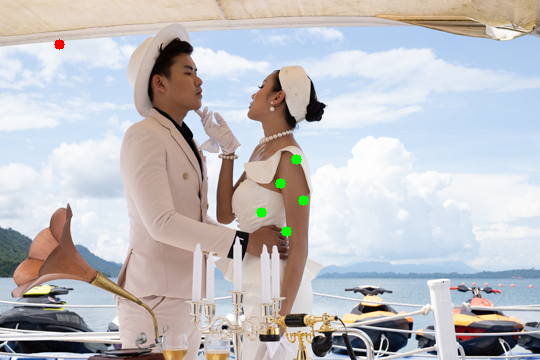}\put(-10,-32){\begin{sideways} \scriptsize{PPR10K-c Dataset}  \end{sideways}}\end{overpic} 
\hspace{-3pt}\includegraphics[width=85pt]{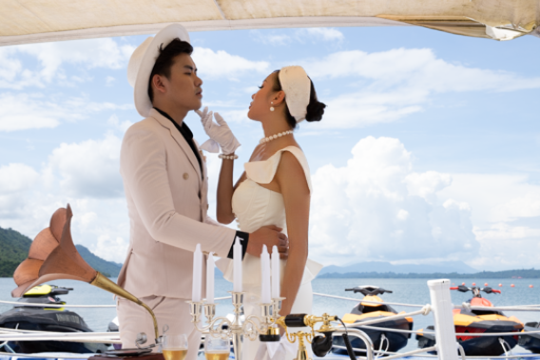}
\includegraphics[width=85pt]{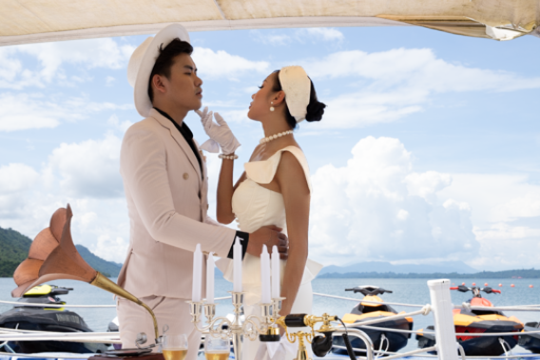}
\includegraphics[width=85pt]{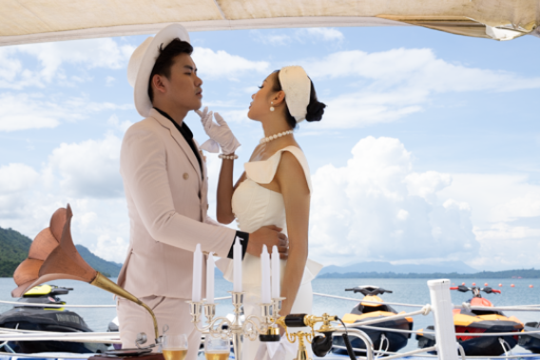}
\includegraphics[width=85pt]{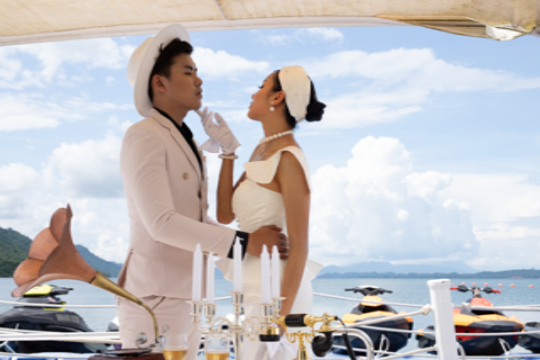}
\includegraphics[width=85pt]{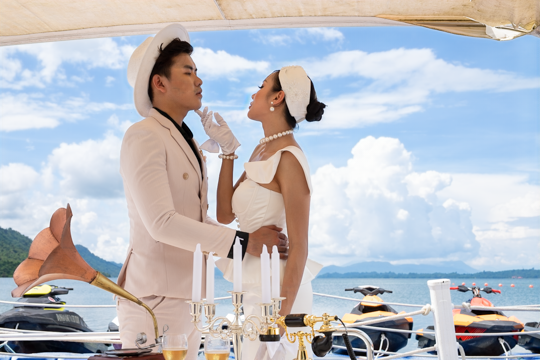} \\

\hspace{1pt}\scriptsize{4.0996} 
\hspace{63pt}\scriptsize{4.5304}
\hspace{63pt}\scriptsize{4.4196}
\hspace{63pt}\scriptsize{4.4333}
\hspace{63pt}\scriptsize{4.9138}
\hspace{63pt}\scriptsize{3.6866}\vspace{1pt}

\includegraphics[width=85pt]{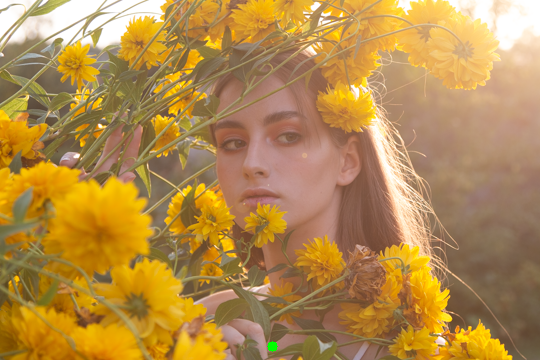}
\includegraphics[width=85pt]{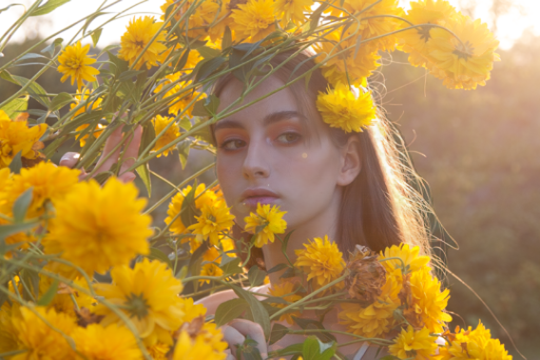}
\includegraphics[width=85pt]{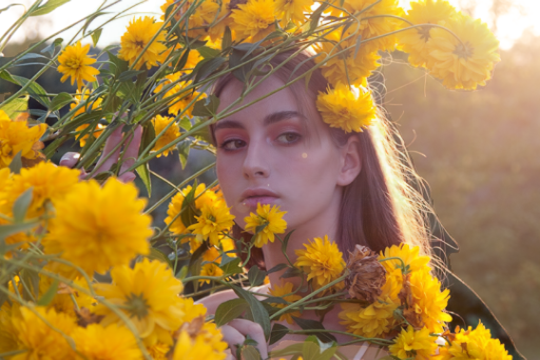}
\includegraphics[width=85pt]{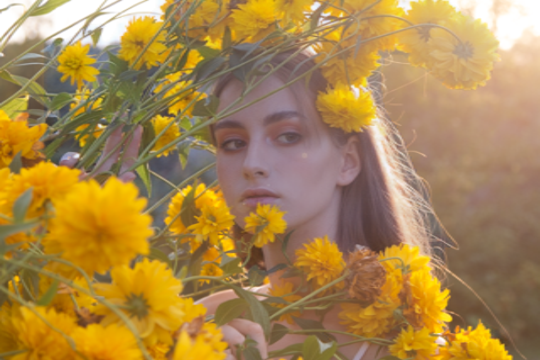}
\includegraphics[width=85pt]{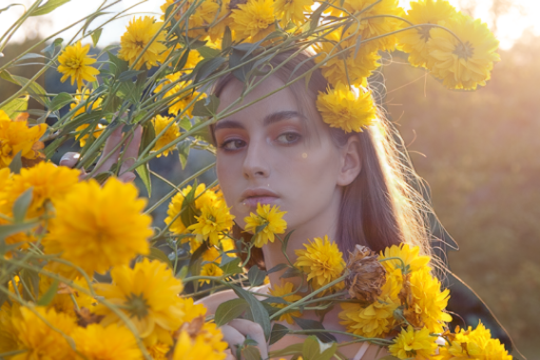}
\includegraphics[width=85pt]{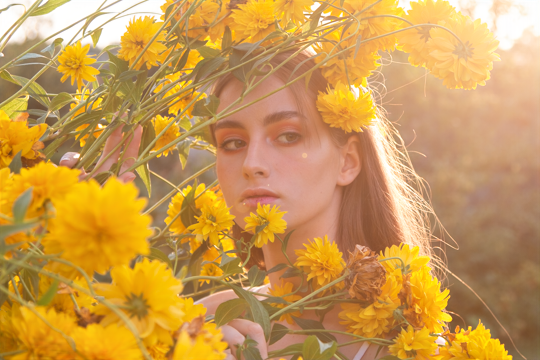} 

\hspace{1pt}\scriptsize{3.3923} 
\hspace{63pt}\scriptsize{3.8660}
\hspace{63pt}\scriptsize{3.6350}
\hspace{63pt}\scriptsize{3.9671}
\hspace{63pt}\scriptsize{3.7334}
\hspace{63pt}\scriptsize{3.3004}\vspace{1pt}

\hspace{-2pt}\scriptsize{(a)~Input}
\hspace{52pt}\scriptsize{(b)~HDRNet}
\hspace{47pt}\scriptsize{(c)~CSRNet}
\hspace{50pt}\scriptsize{(d)~3D LUT}
\hspace{40pt}\scriptsize{(e)~3D LUT HRP}
\hspace{44pt}\scriptsize{(f)~Ours} 

\vspace{-4pt}\caption{Visual comparison of interactive portrait retouching on the PPR10K-a dataset, PPR10K-b dataset and PPR10K-c dataset. \red{The positive/negative clicks are marked with green/red dots to denote the emphasized and non-emphasized retouching regions, respectively.  The number of positive/negative clicks is limited to 5.} All baseline methods are modified with cascading strategy of interactive segmentation-retouching-harmonization, where we set RITM-H18~\cite{sofiiuk2021reviving} and DHT~\cite{guo2021image} to complete the interactive segmentation and image harmonization. We provide the NIQE metric of these images at the bottom.
} 
\label{figure interactive_comparison}
\end{figure*}

\textbf{Quantitative Comparison.}
The non-reference image quality assessment results are reported in Table~\ref{table: user study}, where we mark the best result in bold. It can be observed that the proposed method archives a landslide superiority over other methods, demonstrating that such a naive cascaded strategy design cannot meet the requirement of naturalness and achieve perceptual satisfaction. 

\textbf{Visual Comparison.}
We show the visual comparison for interactive retouching in Fig.~\ref{figure interactive_comparison}, where the baseline methods are modified with the cascading strategy demonstrated in Fig.~\ref{fig:teaser}(b). We can observe that these cascaded interactive baselines fail to keep a balance between the emphasized regions and the background, inevitably leading to artifacts and unnaturalness. 
\pink{While our interactive portrait retouching branch particularly emphasizes user-specified regions and simultaneously retouches the background, which satisfies the requirements of instance adaptivity and naturalness, respectively.} By comparing our interactive results with cascade baseline methods, we aim to demonstrate that the cascaded solution of  \textit{``interactive segmentation - local retouching - image harmonization"} is suboptimal for the interactive retouching task. 

\begin{figure*}[t]
\vspace{-6pt}
\hspace{-2pt}\includegraphics[width=85pt]{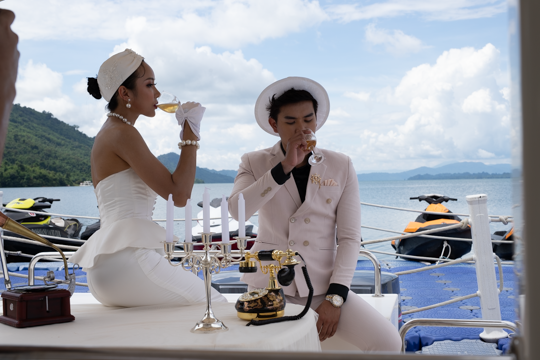}\hspace{-1pt}  
\includegraphics[width=85pt]{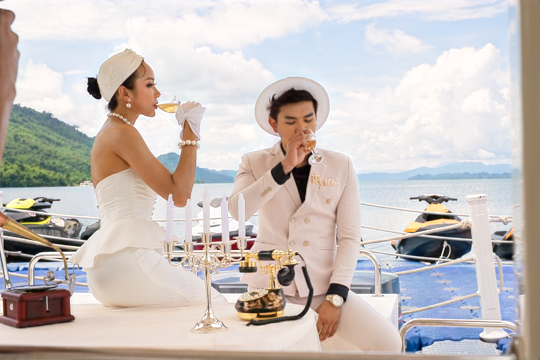}\hspace{-1pt}  
\includegraphics[width=85pt]{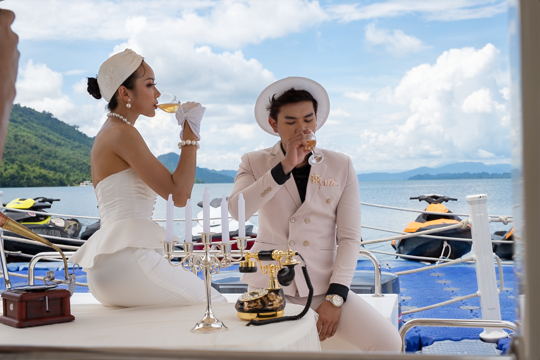}\hspace{-1pt}  
\includegraphics[width=85pt]{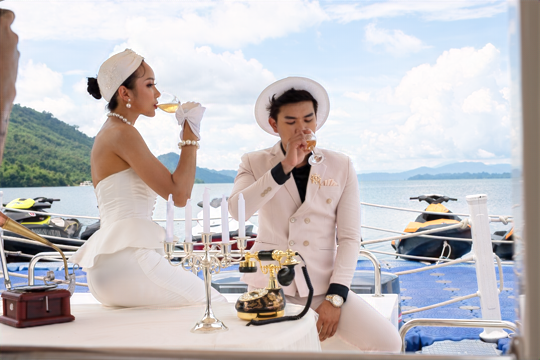}\hspace{-1pt}  
\includegraphics[width=85pt]{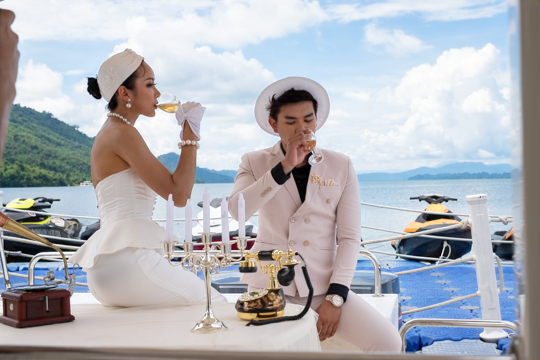}\hspace{-1pt}  
\includegraphics[width=85pt]{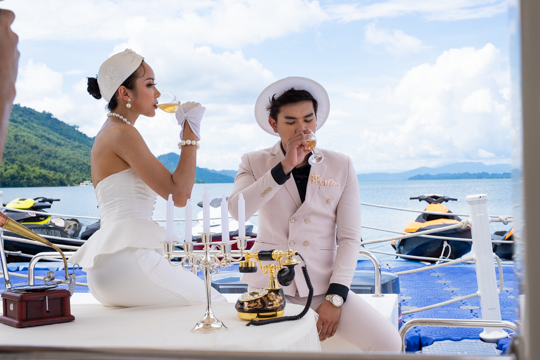}\vspace{2pt}  

\hspace{-2pt}\includegraphics[width=85pt]{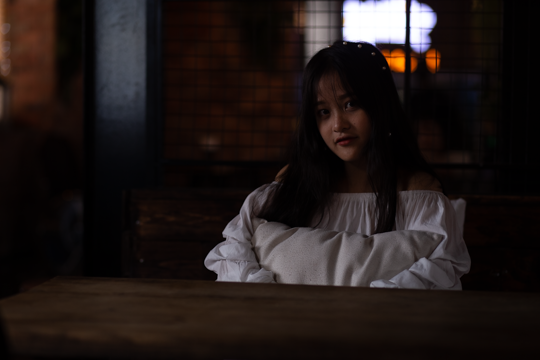}\hspace{-1pt} 
\includegraphics[width=85pt]{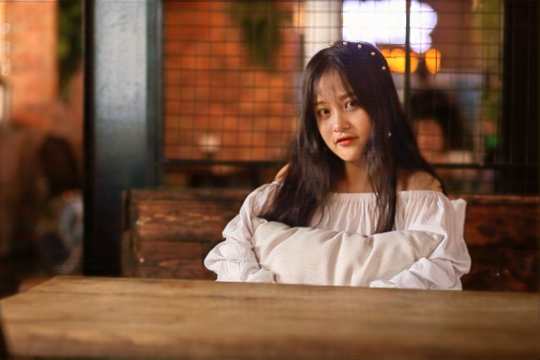}\hspace{-1pt} 
\includegraphics[width=85pt]{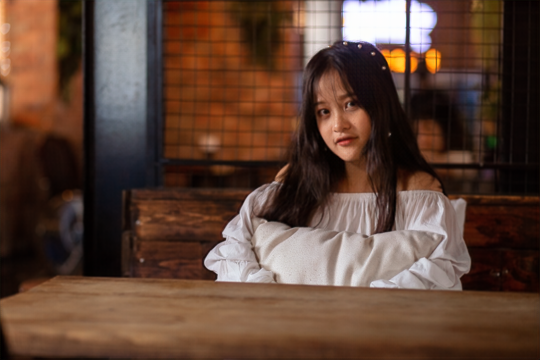}\hspace{-1pt} 
\includegraphics[width=85pt]{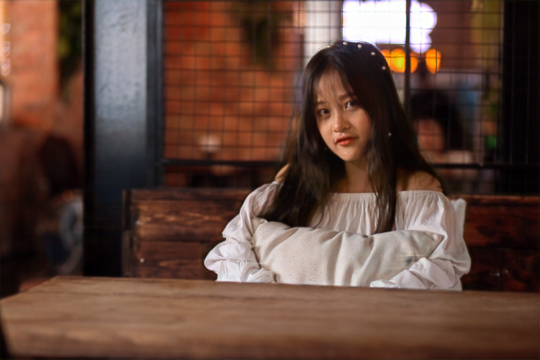}\hspace{-1pt} 
\includegraphics[width=85pt]{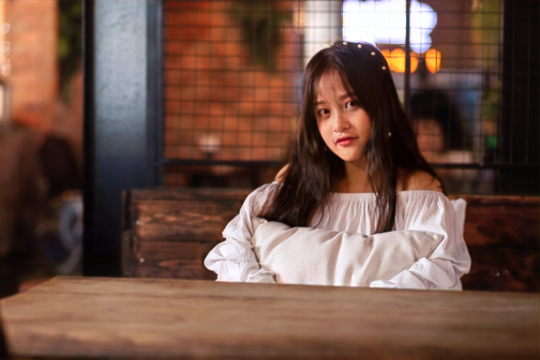}\hspace{-1pt} 
\includegraphics[width=85pt]{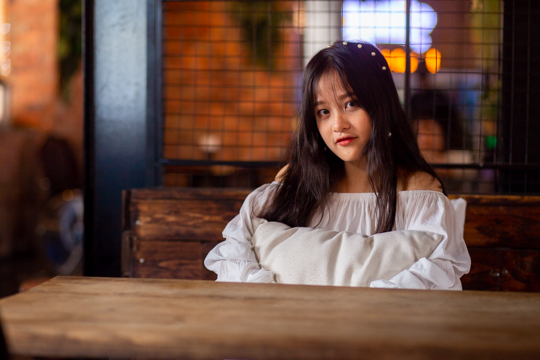}\vspace{-3pt}   

\hspace{-2pt}\subfigure[Input]{\includegraphics[width=85pt, clip=true, trim=0 230pt 0pt  80pt]{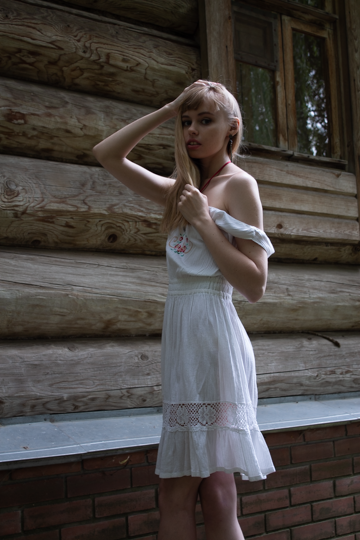}}\hspace{-1pt} 
\subfigure[w/o $E_r$]{\includegraphics[width=85pt, clip=true, trim=0 230pt 0pt  80pt]{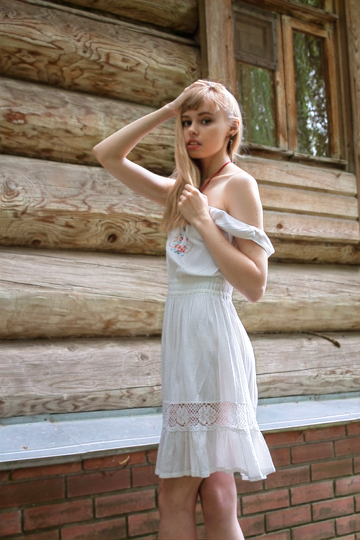}}\hspace{-1pt} 
\subfigure[w/o RF]{\includegraphics[width=85pt, clip=true, trim=0 230pt 0pt  80pt]{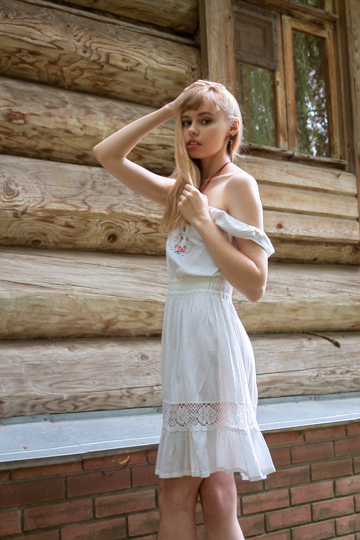}}\hspace{-1pt} 
\subfigure[w/o HRP]{\includegraphics[width=85pt, clip=true, trim=0 230pt 0pt  80pt]{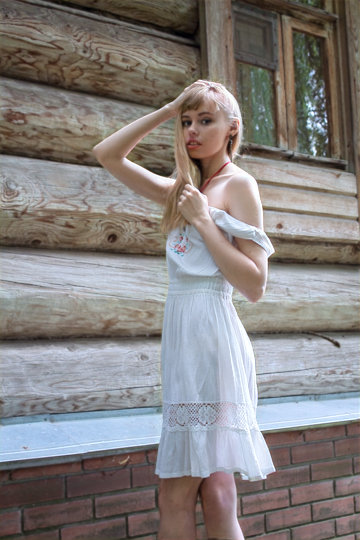}}\hspace{-1pt} 
\subfigure[Ours]{\includegraphics[width=85pt, clip=true, trim=0 230pt 0pt  80pt]{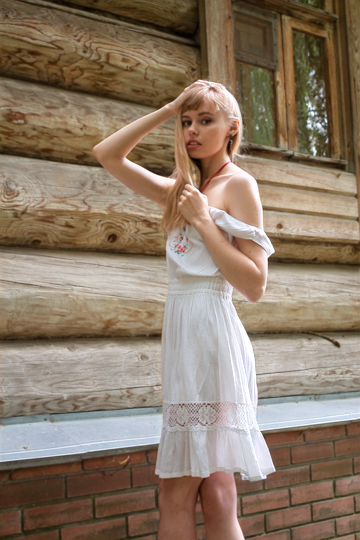}}\hspace{-1pt} 
\subfigure[GT]{\includegraphics[width=85pt, clip=true, trim=0 230pt 0pt  80pt]{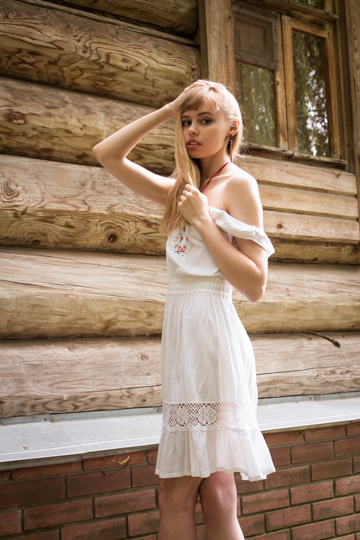}} 

\vspace{-4pt}\caption{Visualization of the ablation study of the automatic region-aware retouching branch, where $E_r$, RF and HRP represents the potential region extractor, region filtering and human-region priority strategy, respectively.}\vspace{-4pt}   

\label{ablation visualization}
\end{figure*}

\subsection{Ablation Study}
In this section, we verify the effectiveness of our method from the architecture design of the automatic retouching branch and the interactive retouching branch.
For the automatic retouching branch, we mainly investigate the following components: the potential region extractor $E_r$, region filtering and human-region priority training strategy.  The results are summarized with PSNR, $\Delta E_{a b}$, $PSNR^{HC}$ and $\Delta E_{a b}^{HC}$.  For the interactive retouching branch, we investigate the effectiveness of the proposed region selection module by comparing it with the concatenation operation. \pink{The results are summarized with the non-reference image quality assessments NIQE, ILNIQE, BRISQUE, MA and BIQI.} All experiments are performed on the PPR10K-c dataset.

\begin{table}[t]
\centering
\caption{Ablation study for the automatic region-aware retouching branch, where $E_r$, RF and HRP stand for potential region extractor $E_r$, region filtering and human-region priority training strategy, respectively.}\vspace{-4pt}  
\label{automatic_ablation} 
 \setlength{\tabcolsep}{1.5mm}{
\begin{tabular}{c c c | c  c c c}
\toprule
 $E_r$ & RF  & HRP & PSNR $\uparrow$ &  $\Delta E_{a b} \downarrow$ & PSNR\small{$^{HC}$} $\uparrow$ & $\Delta E_{a b}$\small{$^{HC}$} $\downarrow$      \\  
\hline

- & $\checkmark$ & $\checkmark$ & 25.33& 	8.06&  28.65& 	5.19 \\
$\checkmark$ &-&  $\checkmark$  &  25.24& 	7.64 &  28.54& 	4.95  \\
$\checkmark$ &  $\checkmark$ & - & 25.38& 	7.73 & 28.65& 	5.01   \\
$\checkmark$ & $\checkmark$ & $\checkmark$ & \textbf{25.68}& 	\textbf{7.41} & \textbf{28.97}& 	\textbf{4.80} \\
\bottomrule
\end{tabular}
}\vspace{-16pt}
\end{table}

\textbf{Potential Region Extractor $\boldsymbol{E_r}$.} As mentioned in Sec.~\ref{automatic retouching}, we utilize the potential region extractor $E_r$ to search regions of interest and provide region candidates for the retouching network. Here, we investigate the role of $E_r$ by removing it, and verify whether the automatic retouching branch can handle region-aware retouching alone. As shown in Table~\ref{automatic_ablation}, we can observe that without extracting features of potential regions, the retouching network cannot adequately deal with automatic region-aware retouching.  \pink{In addition, it leads to the worst results on CIELAB based assessments $\Delta E_{a b}$ and $\Delta E_{a b}^{HC}$ among all ablation settings}, demonstrating the importance of the potential region extractor $E_r$.  The results displayed in Fig.~\ref{ablation visualization}(b) show global retouching instead of emphasizing the retouching of human-regions.  We suppose this is because without the help of $E_r$, the retouching network needs to search regions of interest and perform retouching simultaneously, thus leading to the performance drop of automatic retouching.

\textbf{Region Filtering.  } We predict a soft mask at the end of the decoding process to filter out human-irrelevant regions and impose $\mathcal{L}_{mask}$ on it.  To verify the effectiveness of region filtering, we remove the soft mask $\tilde{M}$ prediction process and discard the element-wise multiplication between $f_y$ and $\tilde{M}$ at the same time.  As listed in Table~\ref{automatic_ablation}, there is a noticeable performance drop especially for the human-centered assessments $PSNR^{HC}$ and $\Delta E_{a b}^{HC}$, demonstrating that it is essential to apply region filtering to filter out human-irrelevant regions.   \red{The visual results provided in Fig.~\ref{ablation visualization}(c) also show that the automatic retouching branch fails to emphasize human-region retouching without region filtering.}

\textbf{Human-region Priority Training Strategy.}
Our method imposes the human-region weighted L1 loss $\mathcal{L}_{prior}$ (see Eq.~\ref{retouching}) following the human-region priority (HRP) strategy~\cite{liang2021ppr10k}. To investigate the effectiveness of HRP, we set the human-region weight $\boldsymbol{W}$ to 1 so that all pixels are treated equally. The results in Table~\ref{automatic_ablation} show a significant performance drop and Fig.~\ref{ablation visualization}(d) shows that without the HRP strategy, the model fails to pay special attention on human-region and leads to flat-looking results.

 \begin{table}[t]
  \centering
\caption{Ablation study for the interactive region-ware retouching branch. The proposed region selection module shows a clear advantage over straight-forward concatenation solution.}\vspace{-4pt}
  \label{interactive ablation}
    \setlength{\tabcolsep}{0.85mm}{
  \begin{tabular}{c| c c c c c c}
    \toprule
     Fusion type   & NIQE $\downarrow$ &  ILNIQE $\downarrow$ & BRISQUE $\downarrow$ & MA $\uparrow$   & BIQI $\downarrow$    \\  
   \hline
     Concatenation   & 3.7794 &  25.0494 &  19.7681	 &  8.2335 & 28.5029     \\
\scriptsize{Region Selection Module}   & \textbf{3.7768} & \textbf{24.9507}  &  \textbf{19.4812}   & \textbf{8.2514} & \textbf{28.2272} \\ 
    \bottomrule
  \end{tabular} 
  }
\end{table}

\begin{figure}[t]

 \hspace{6pt}\begin{overpic}[width=81pt]{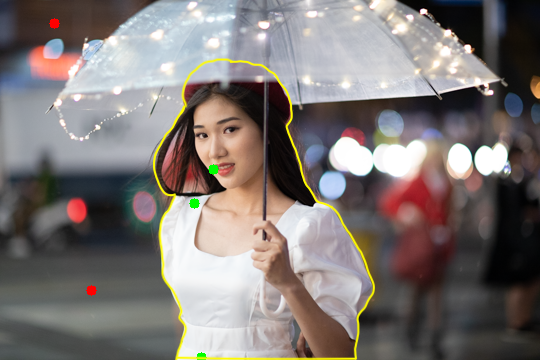}\put(-10,9){\begin{sideways} \scriptsize{Concatenation} \end{sideways}}\end{overpic}\hspace{-1pt} 
\includegraphics[width=81pt]{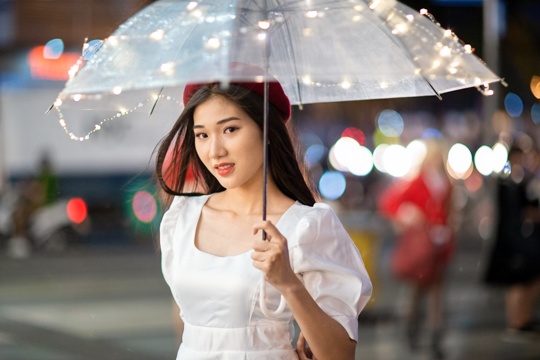}\hspace{-1pt} 
\includegraphics[width=81pt]{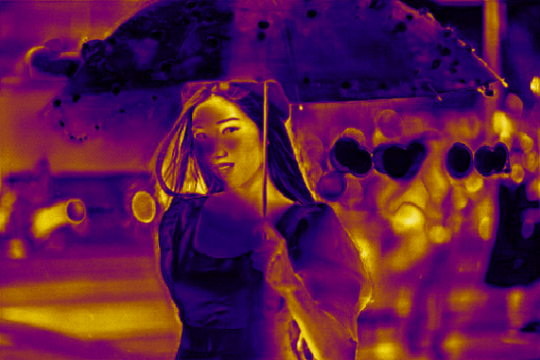} 

\hspace{36pt}\scriptsize{4.3968} 
\hspace{64pt}\scriptsize{4.1455} 
\vspace{2pt} 

 \hspace{6pt}\begin{overpic}[width=81pt]{Figures/ablation/cat_our_ablation/cont_click_1621_4.png}\put(-10,13){\begin{sideways} \scriptsize{RS Module}  \end{sideways}}\end{overpic}\hspace{-1pt} 
\includegraphics[width=81pt]{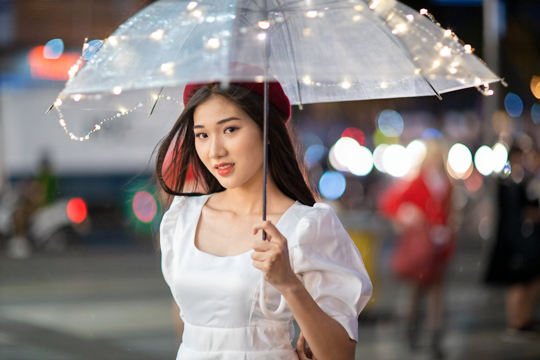}\hspace{-1pt}  
\includegraphics[width=81pt]{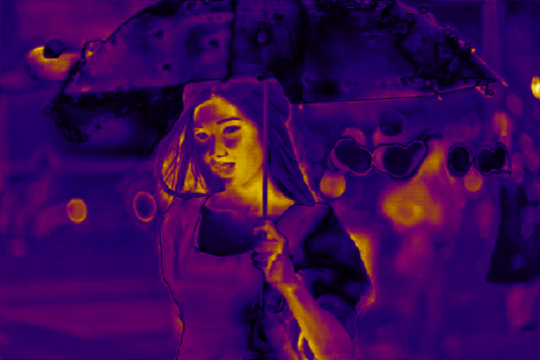} 

\hspace{36pt}\scriptsize{4.3968} 
\hspace{64pt}\scriptsize{4.0193}
\vspace{2pt} 

 \hspace{6pt}\begin{overpic}[width=81pt,clip=true, trim=0 275pt 0  20pt]{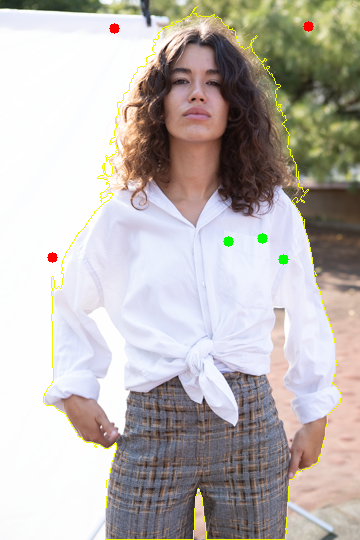}\put(-10,9){\begin{sideways} \scriptsize{Concatenation}  \end{sideways}}\end{overpic}\hspace{-1pt} 
\includegraphics[width=81pt,clip=true, trim=0 275pt 0  20pt]{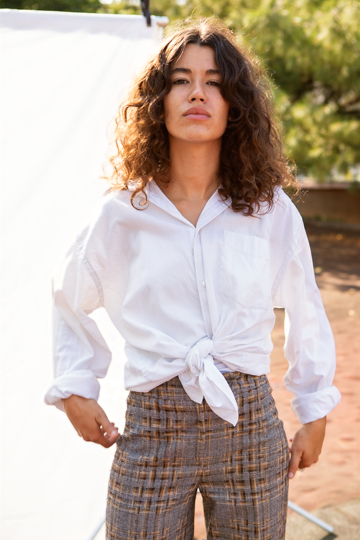}\hspace{-1pt}
\includegraphics[width=81pt,clip=true, trim=0 275pt 0  20pt]{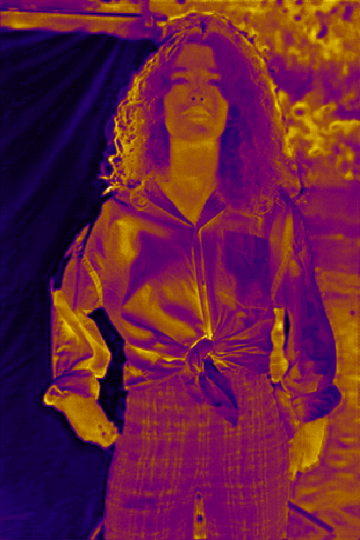} 

\hspace{36pt}\scriptsize{4.1487} 
\hspace{64pt}\scriptsize{3.1862} 
\vspace{2pt} 

 \hspace{6pt}\begin{overpic}[width=81pt,clip=true, trim=0 275pt 0  20pt]{Figures/ablation/cat_our_ablation/cont_click_1437_1.png}\put(-10,13){\begin{sideways} \scriptsize{RS Module}  \end{sideways}}\end{overpic}\hspace{-1pt} 
\includegraphics[width=81pt,clip=true, trim=0 275pt 0  20pt]{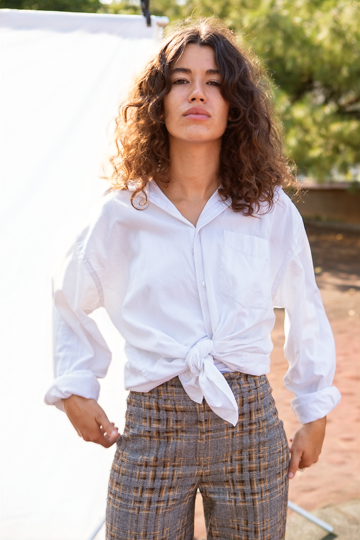}\hspace{-1pt} 
\includegraphics[width=81pt,clip=true, trim=0 275pt 0  20pt]{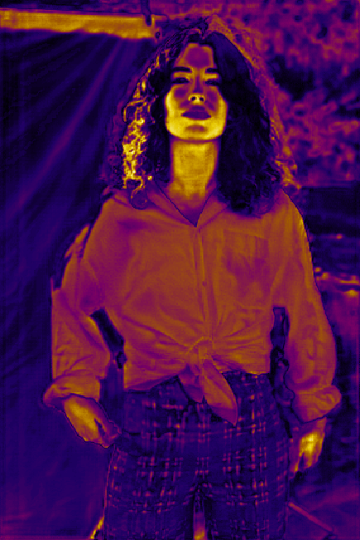}\hspace{-1pt}  

\hspace{36pt}\scriptsize{4.1487} 
\hspace{64pt}\scriptsize{3.1225}
\vspace{2pt} 

\hspace{29pt}\footnotesize{(a)~Input}
\hspace{51pt}\footnotesize{(b)~Output}
\hspace{36pt}\footnotesize{\rose{(c)~Residual Map}}
\vspace{-4pt}\caption{Visualization of the ablation study of the interactive region-aware retouching branch, where RS Module is short for the Region Selection Module. The user-given positive/negative clicks are marked with green/red dots. We distinguish the emphasized and non-emphasized portrait regions with a yellow curve. \rose{For each image, we provide the NIQE metric at the bottom, and include a residual map to reflect the pixel change between input and output.} }\vspace{-0.1in} 
\label{fig: ablation for interactive branch}
\end{figure}

\textbf{Region Selection Module. } To demonstrate the effectiveness of our region selection module, we replace the region selection module with simple concatenation to fuse the encoded priority condition vector $z$ and region-aware features $f_r$.  As depicted in Table~\ref{interactive ablation}, the proposed region selection module performs better in generating natural results.   \rose{The visual comparisons are shown in Fig.~\ref{fig: ablation for interactive branch}.  The user-given positive/negative clicks are marked with green/red dots to denote the emphasized and non-emphasized regions (also distinguished by the yellow curve).  To intuitively compare the visual quality of the proposed region selection module and concatenation operation, we include a residual map for each image.  The residual maps are generated by measuring the L1 distance between the inputs and results, indicating the pixel change introduced by the interactive retouching branch. The higher residual value denotes more attention is paid to retouching this region.  As shown in Fig.~\ref{fig: ablation for interactive branch}, the region selection module effectively captures users’ intents and associates sparse user guidance with image pixels (e.g., the human region is emphasized and the background is non-emphasized, while the concatenation-based implementation simultaneously emphasizes human-region and the background).}

\subsection{Beyond Portrait Retouching}
\begin{figure}[t]
\hspace{6pt}\begin{overpic}[width=81pt]{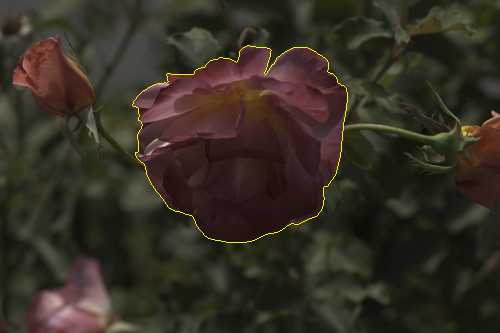}\put(-10,15){\begin{sideways} \scriptsize{Automatic}  \end{sideways}}\end{overpic}\hspace{-1pt}  
\includegraphics[width=81pt]{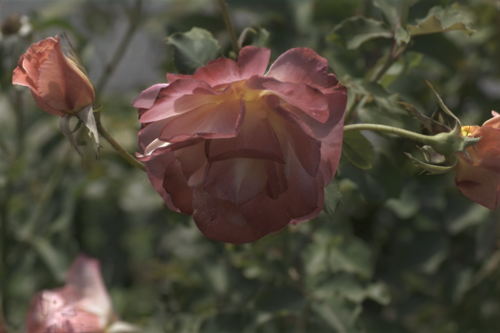}\hspace{-1pt}  
\includegraphics[width=81pt]{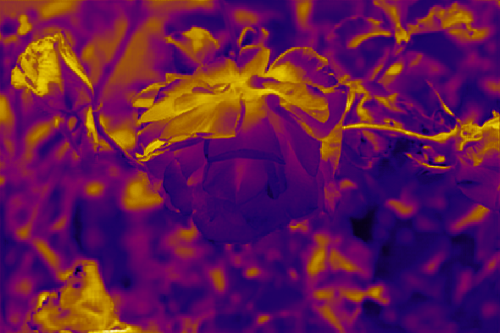}\vspace{1pt} 

\hspace{6pt}\begin{overpic}[width=81pt]{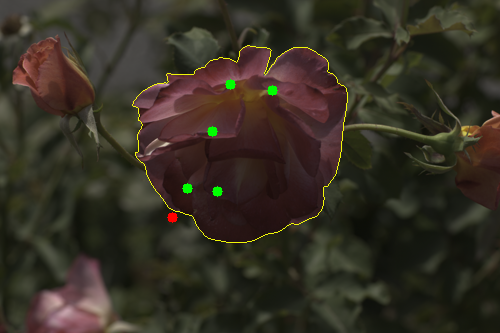}\put(-10,15){\begin{sideways} \scriptsize{Interactive}  \end{sideways}}\end{overpic}\hspace{-1pt} 
\includegraphics[width=81pt]{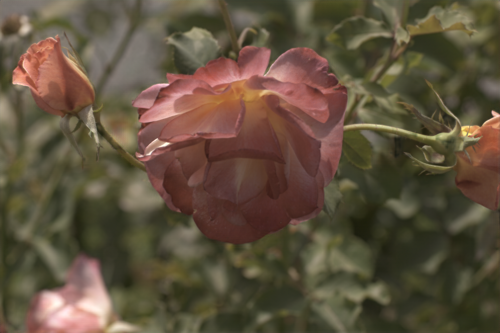}\hspace{-1pt} 
\includegraphics[width=81pt]{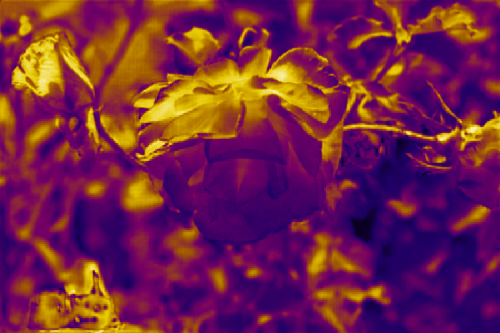}\vspace{4pt}

\hspace{6pt}\begin{overpic}[width=81pt]{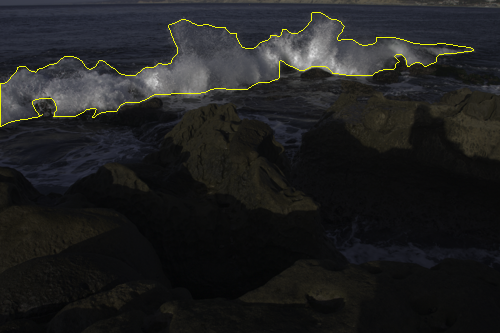}\put(-10,15){\begin{sideways} \scriptsize{Automatic}  \end{sideways}}\end{overpic}\hspace{-1pt} 
\includegraphics[width=81pt]{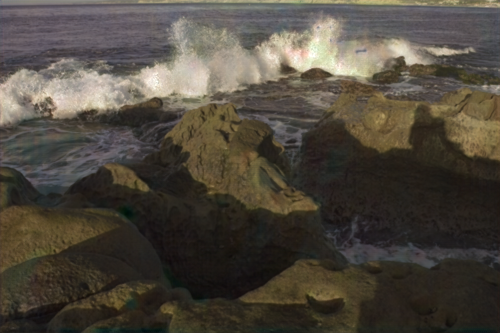}\hspace{-1pt} 
\includegraphics[width=81pt]{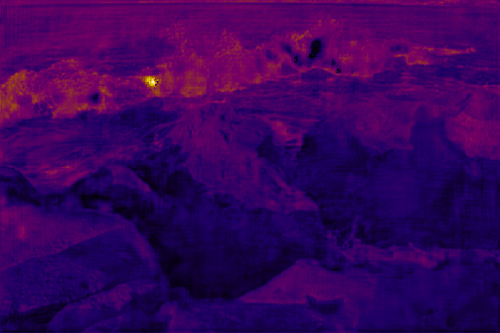}\vspace{1pt}

\hspace{6pt}\begin{overpic}[width=81pt]{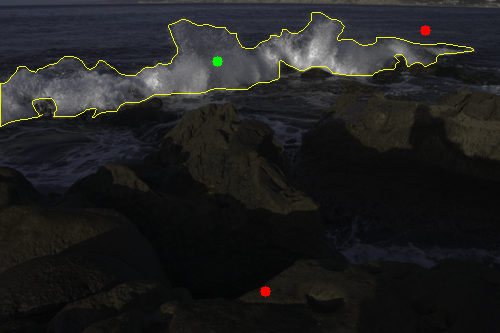}\put(-10,15){\begin{sideways} \scriptsize{Interactive}  \end{sideways}}\end{overpic}\hspace{-1pt} 
\includegraphics[width=81pt]{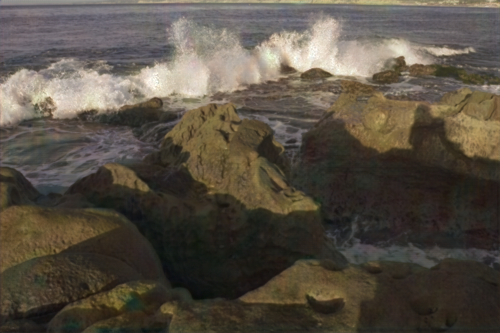}\hspace{-1pt}  
\includegraphics[width=81pt]{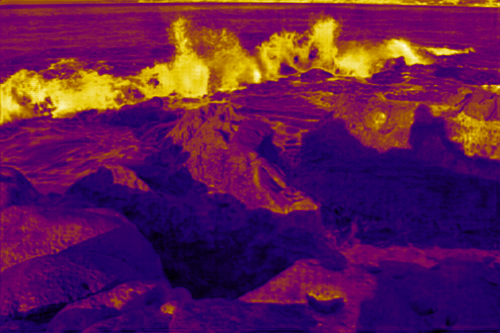}\vspace{4pt}

\hspace{6pt}\begin{overpic}[width=81pt]{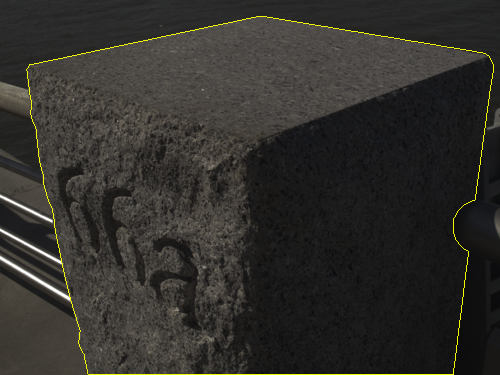}\put(-10,20){\begin{sideways} \scriptsize{Automatic}  \end{sideways}}\end{overpic}\hspace{-1pt} 
\includegraphics[width=81pt]{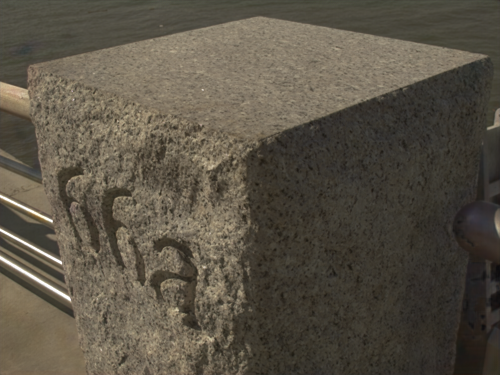}\hspace{-1pt} 
\includegraphics[width=81pt]{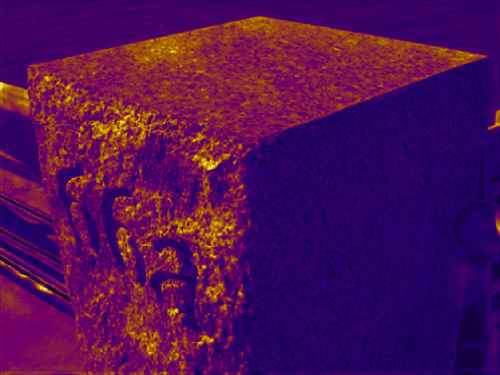}\vspace{2pt}

\hspace{6pt}\begin{overpic}[width=81pt]{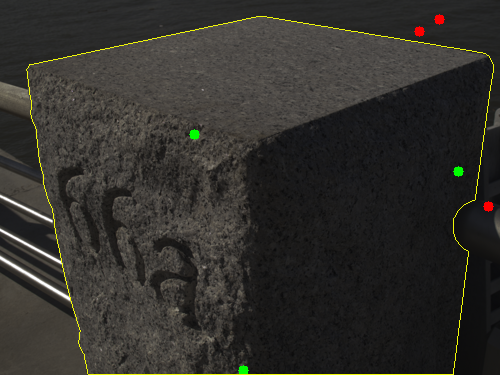}
\put(-10,20){\begin{sideways} \scriptsize{Interactive}  \end{sideways}}
\put(32,-14){\footnotesize{(a) Input}}
\end{overpic}\hspace{-1pt} 
\begin{overpic}[width=81pt]{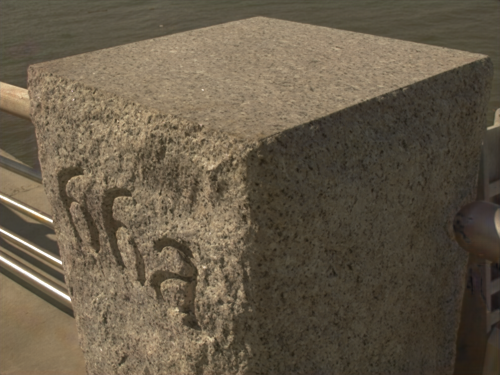}
\put(28,-14){\footnotesize{(b) Output}}
\end{overpic}\hspace{-1pt} 
\begin{overpic}[width=81pt]{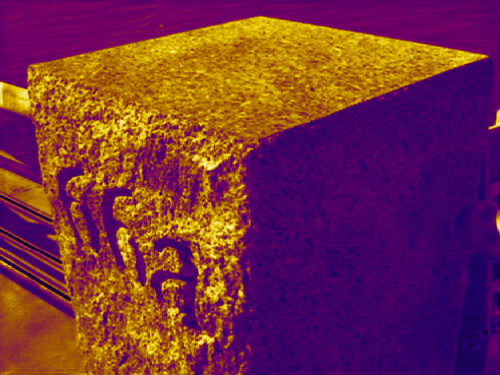}
\put(16,-14){\rose{\footnotesize{(c) Residual Map}}}
\end{overpic}\vspace{10pt} 

\caption{Results of the generalization ability test on the MIT-Adobe FiveK dataset, where green/red dots denote the positive/negative clicks.  We mark the focal regions with yellow curves.  \rose{A residual map is provided for each image to intuitively demonstrate the improvement introduced by user guidance.  Compared with the automatic branch, our interactive branch successfully emphasizes the user-specified region and shows generalization ability on other non-portrait scenes,} demonstrating its potential ability for more applications.  }\vspace{-10pt}
\label{fivek}
\end{figure}

To test the generalization ability of our interactive retouching on other scenes, we apply the trained models to samples from the MIT-Adobe FiveK dataset~\cite{mitfivek} (including many non-portrait scenes) without fine-tuning.  As shown in Fig.~\ref{fivek},  we mark the emphasized region with the yellow curve. \rose{Residual maps that compute the $L_1$ loss between the inputs and results are provided to intuitively show the difference.} For the interactive retouching, the positive and negative clicks are denoted with green and red dots, respectively. \rose{It can be observed that based on the automatic branch, residual maps of the interactive branch contain more pixel changes in the emphasized region, indicating that our interactive branch adequately utilizes users' intents and emphasizes the retouching of user-specified regions (e.g., the flower of the first scene).} By fully exploring users' intent, our interactive branch successfully handles unseen scenes and classes (e.g., \red{the waves and the stone}).

\section{Conclusions and Future Work}
In this work, we explore the interactive region-aware portrait retouching task and propose a novel unified framework, which can handle automatic retouching and provide flexibility for interactive retouching.   To effectively capture users' intents and associate the user guidance with relevant regions, we consider the interactive retouching as a latent feature editing task and propose a region selection module to modulate the extracted image semantics under the user guidance.   Experimental results show the ability of our interactive branch to capture users' intents as well as the state-of-the-art performance of our automatic branch.  

The inference of interactive retouching costs approximately 178 ms on a single NVIDIA GTX1060 GPU, which basically satisfies the real-time requirement for interactive tasks but should be further improved.  Besides, we currently only support click-based interactive guidance, and our method would be more general if various user input types were considered.  According to users' feedback, iteratively adjusting the interactive retouching results should also be included.   \red{Considering the real-world scenarios such as low-resolution~\cite{ge2018low,gharbi2017deep}, low-light~\cite{he2020conditional,wang2018gladnet}, and occlusion situations~\cite{ge2020occluded,ge2017detecting}, where the portrait regions are ambiguously defined, interactive user guidance would provide more reliable spatial prior for the retouching model instead of blindly searching and retouching plausible regions.  We leave the aspects above as our future work.}

\bibliographystyle{IEEEtran}
\bibliography{egbib.bib}

\end{document}